\documentclass[]{bytedance_seed}

\usepackage[toc,page,header]{appendix}

\usepackage{minitoc}

\usepackage{microtype}
\usepackage{hyperref}
\usepackage{url}
\usepackage{booktabs}

\usepackage{graphicx}

\usepackage{lineno}

\usepackage{algorithm}
\usepackage{algpseudocode}

\usepackage{makecell}

\usepackage{wrapfig}
\usepackage{overpic}
\usepackage{float}
\usepackage{rotating}

\usepackage{float}
\usepackage{algorithmicx}
\restylefloat{algorithm}  

\usepackage{tipa}

\usepackage{amsmath,bm}
\usepackage{amssymb}
\usepackage{mathtools}
\usepackage{amsthm}
\usepackage{listings}
\usepackage{accents}

\usepackage{multirow}
\usepackage{threeparttable}

\usepackage{tabularx}
\usepackage[table]{xcolor}

\usepackage{caption}
\usepackage{etoolbox}

\usepackage{array}     
\usepackage{colortbl}
\usepackage{xcolor}
\usepackage{multicol}  

\definecolor{damaiblue}{RGB}{0, 102, 204}

\makeatletter
\AfterEndEnvironment{algorithm}{\let\@algcomment\relax}
\AtEndEnvironment{algorithm}{\kern2pt\hrule\relax\vskip3pt\@algcomment}
\let\@algcomment\relax
\newcommand\algcomment[1]{\def\@algcomment{\footnotesize#1}}
\renewcommand\fs@ruled{\def\@fs@cfont{\bfseries}\let\@fs@capt\floatc@ruled
  \def\@fs@pre{\hrule height.8pt depth0pt \kern2pt}%
  \def\@fs@post{}%
  \def\@fs@mid{\kern2pt\hrule\kern2pt}%
  \let\@fs@iftopcapt\iftrue}
\makeatother

\definecolor{darkblue}{rgb}{0, 0, 0.5}
\hypersetup{colorlinks=true, citecolor=darkblue, linkcolor=darkblue, urlcolor=darkblue}

\newcommand{\newtext}[1]{\textcolor{black}{#1}}

\title{Virtual Width Networks}

\affiliation[1]{ByteDance Seed}
\contribution{See Contributions section for full author list.}

\abstract{
We introduce Virtual Width Networks (VWN), a framework that delivers the benefits of wider representations without incurring the quadratic cost of increasing the hidden size. VWN decouples representational width from backbone width, expanding the embedding space while keeping backbone compute nearly constant. In our large‑scale experiment, an 8× expansion accelerates optimization by over 2× for next‑token and 3× for next‑2‑token prediction. The advantage amplifies over training as both the loss gap grows and convergence‑speedup ratio increase, showing that VWN is not only token‑efficient but also increasingly effective with scale. Moreover, we identify an approximately log‑linear scaling relation between virtual width and loss reduction, offering an initial empirical basis and motivation for exploring virtual‑width scaling as a new dimension of large‑model efficiency.
}

\date{\today}
\correspondence{Defa Zhu at \email{zhudefa@bytedance.com}}

\begin{document}
\maketitle


\begin{figure}[h!]
    \centering
    \begin{minipage}{0.33\textwidth}
        \centering
        \includegraphics[width=\linewidth]{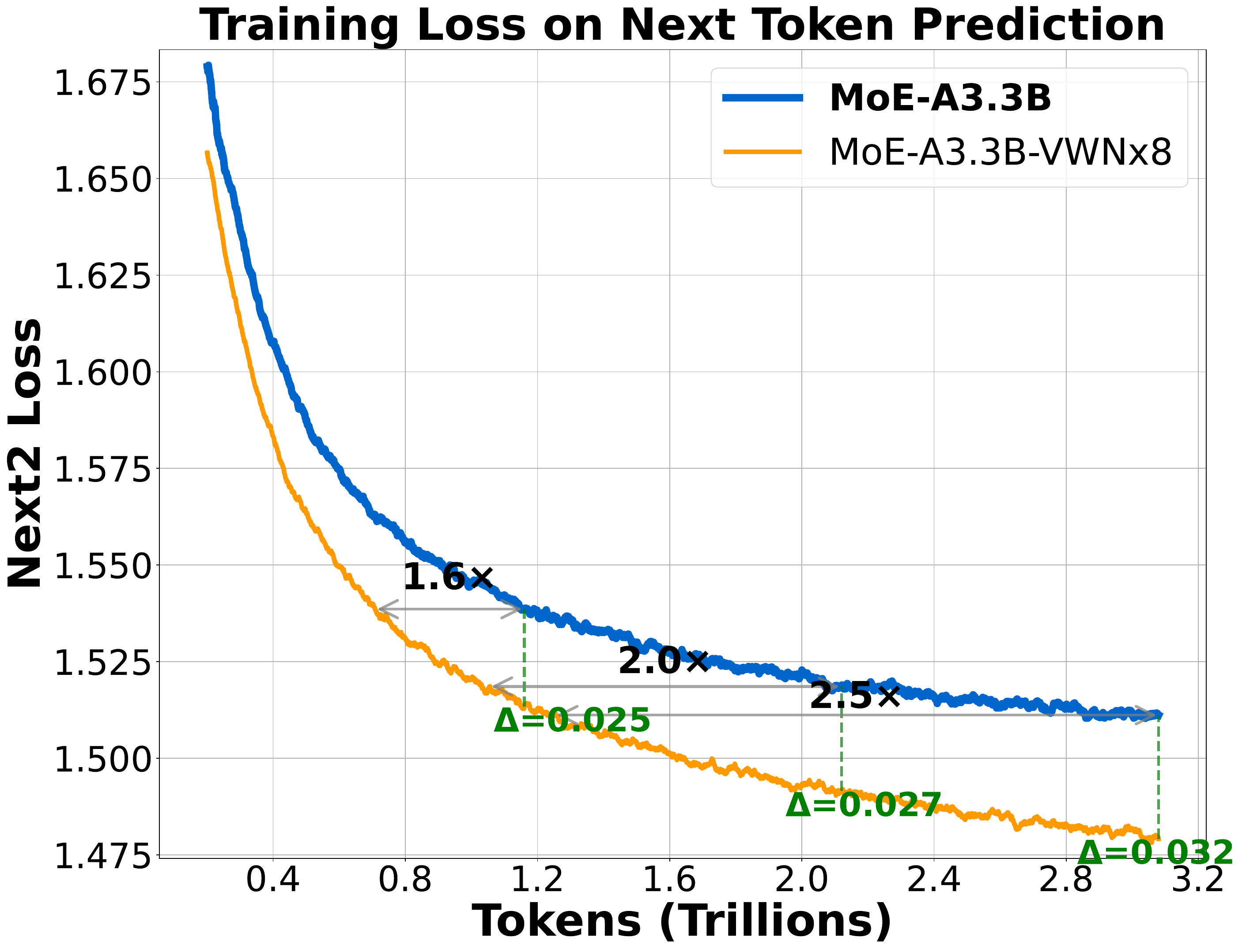}
    \end{minipage}\hfill
    \begin{minipage}{0.33\textwidth}
        \centering
        \includegraphics[width=\linewidth]{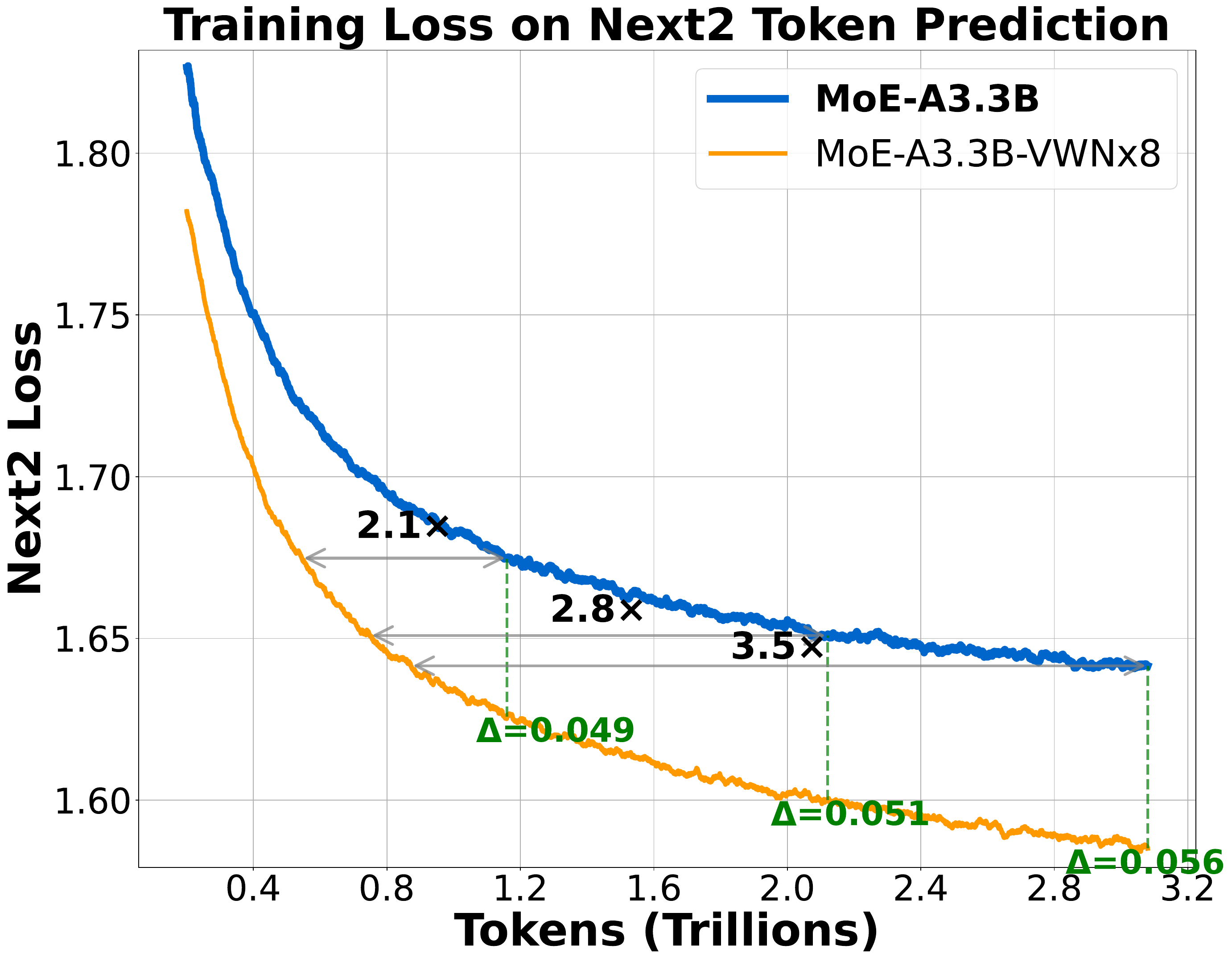}
    \end{minipage}
    \begin{minipage}{0.33\textwidth}
        \centering
        \includegraphics[width=\linewidth]{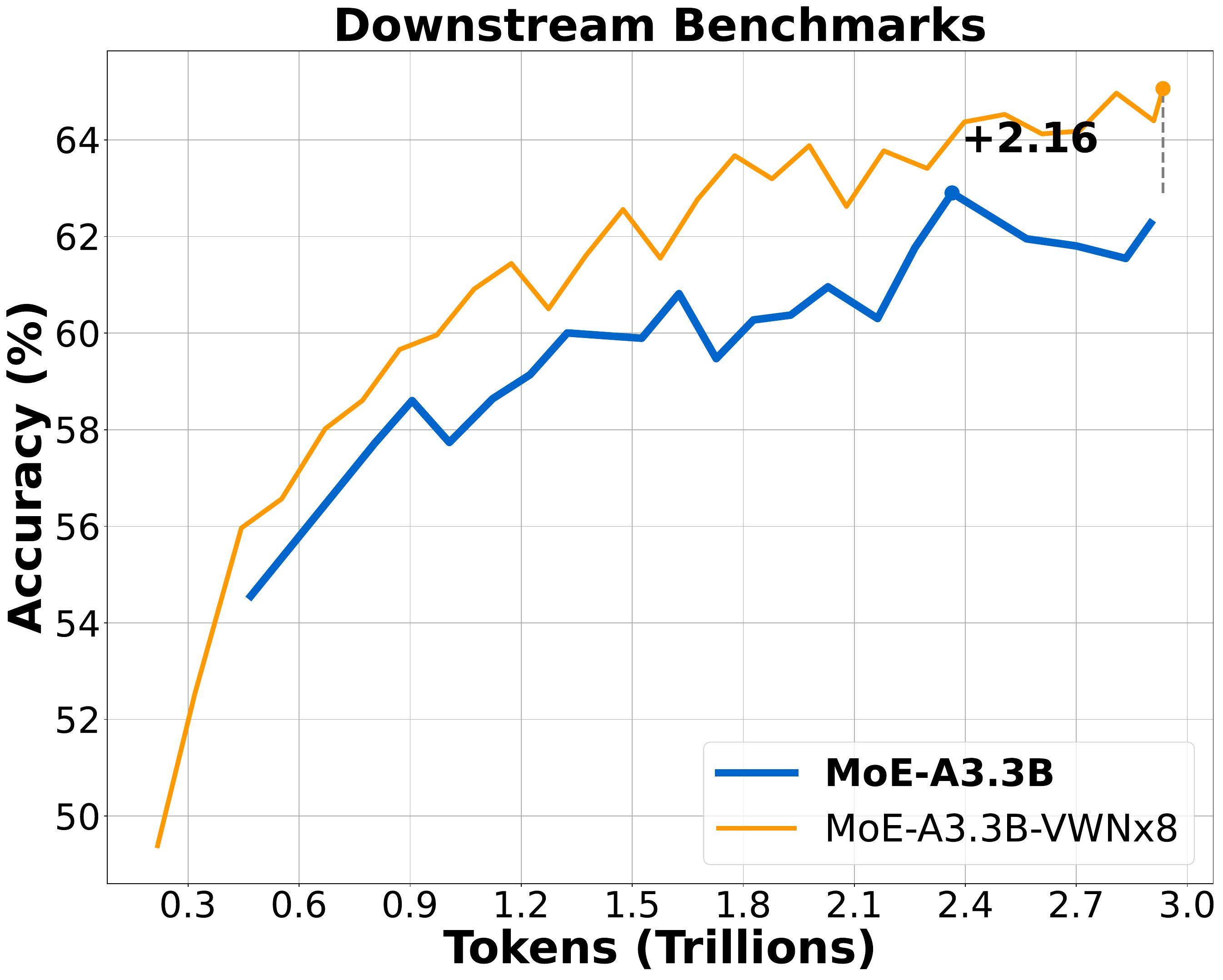}
    \end{minipage}
    \caption{Results from large-scale experiments on a 3.3B-activation MoE using Virtual Width Networks (VWN). We compare the baseline \texttt{MoE-A3.3B} against \texttt{MoE-A3.3B-VWNx8}, configured with a virtual width factor of $r{=}8$. \textbf{Left and middle:} training loss for next-token and next-two-token prediction versus seen tokens. VWN reaches the same loss as the baseline using $2.5\times$ and $3.5\times$ fewer tokens, respectively. \textbf{Right:} average accuracy on a collection of open-source benchmarks (see Table~\ref{tab:benchmarks_b}), where scores are aggregated using internally defined task weights. A difference of one point corresponds to a notable performance gap under this weighting scheme.}
\label{fig:174b}
    \label{fig:174b}
\end{figure}
\section{Introduction}
\begin{figure*}[t]
\centering
\includegraphics[width=0.8\textwidth]{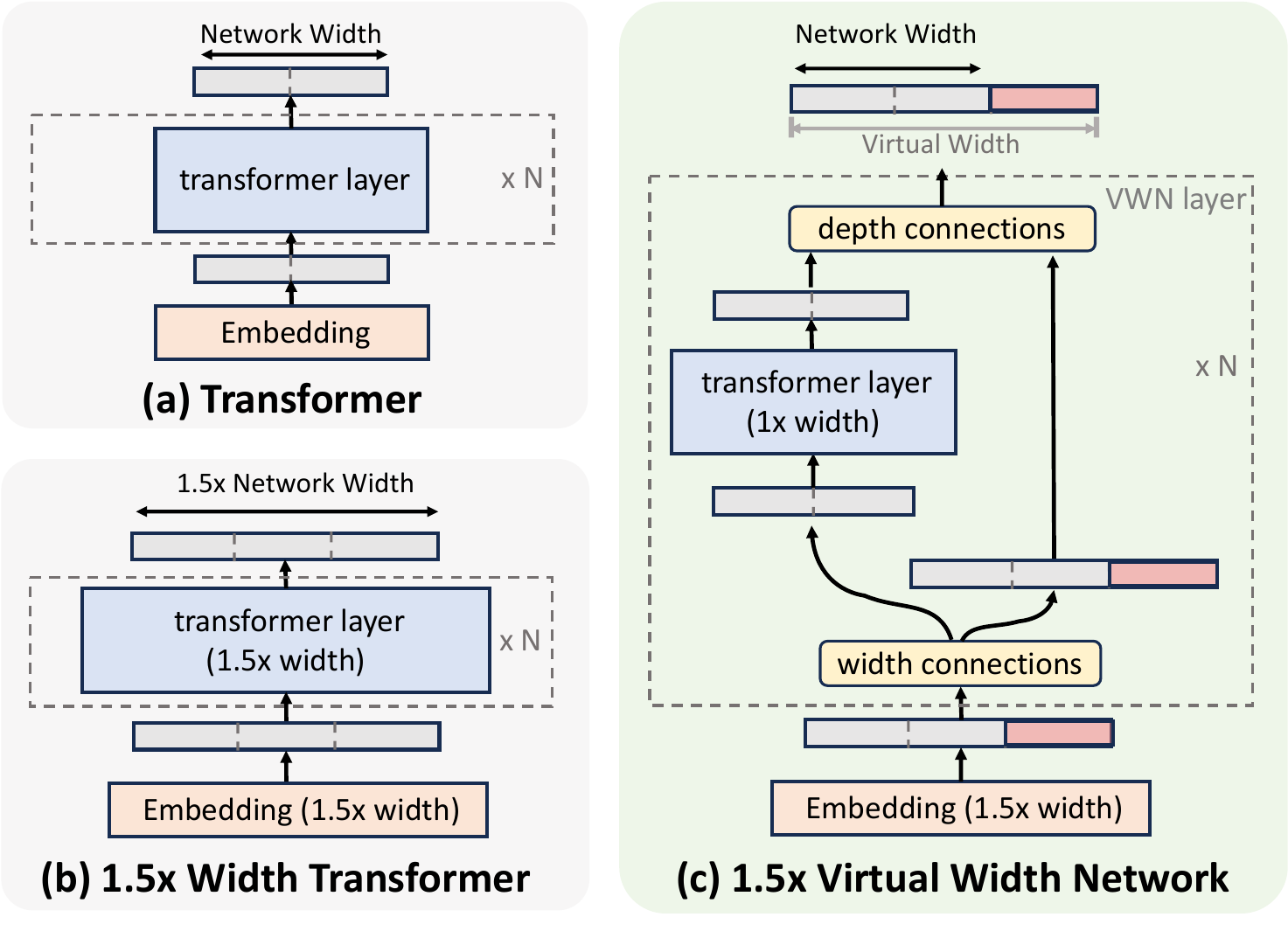}
\caption{\textbf{Standard Transformer vs.\ Virtual Width Network (VWN).}
(a) A standard Transformer uses the same width for embeddings and backbone.
(b) Naive width scaling expands both proportionally, causing quadratic growth in parameters and compute.
(c) \textbf{VWN} decouples embedding width from backbone width. With Generalized Hyper‑Connections, over‑width embeddings (e.g., 1.5$\times$) are coupled to a standard‑width backbone, increasing representational capacity with minimal compute overhead.}
\label{fig:vwn_briefview}
\end{figure*}
According to scaling laws \citep{kaplan2020scaling, hoffmann2022training}, expanding either model parameters or the size of the training corpus yields more capable models.
In particular, increasing the model width (hidden dimensions) enables the representation of richer, more complex functions by packing additional information into each vector, which in turn substantially boosts performance. 
However, naively increasing hidden dimensions leads to quadratic growth in parameters and compute, posing challenges in resource-constrained settings.

To address the challenge and enhance the scalability of modern Transformers, researchers have developed conditional computation strategies that expand model capacity without proportionally increasing computational costs. A prominent example is mixture-of-experts (MoE) architectures \citep{shazeer2017sparsely, lepikhin2020gshard, fedus2022switch}, which dynamically activate specialized subnetworks per input token. By selectively engaging only a fraction of the total parameters for each token during computation, MoE models significantly improve throughput and enable efficient scaling to very large model sizes, without proportionally increasing per-token computational cost.

However, conventional MoE architectures can be viewed as expanding only the inner dimension of the feed-forward networks (FFN), while the backbone hidden dimension remains fixed. Consequently, the model’s representational capacity is still bottlenecked by the hidden dimension, resulting in a persistent performance gap compared to models with truly wider hidden layers. While directly increasing the hidden dimension can close this gap, it incurs a quadratic increase in parameters and computation. This prompts us to ask: \textit{Can we harness the benefits of wider representations while avoiding the quadratic cost explosion of naive scaling?}


In this work, we address this challenge by proposing Virtual Width Networks (VWN), a general framework that enables scaling token‑embedding width while keeping the hidden dimensions of the Transformer backbone fixed. Our key insight is that wider representations can be achieved by expanding the embeddings rather than by widening the hidden layers, the latter being the main source of quadratic computational cost. In this view, models employing methods such as Hyper‑Connections~\cite{zhu2024hyper} or AltUp~\citep{baykal2023alternating} can be regarded as simplified instances within the broader VWN family. By enhancing the design of VWN, we further improve its representational capacity and uncover a favorable scaling property of virtual width—specifically, a scaling relation between the loss and the virtual width factor under a fixed backbone—which offers the community a new dimension for scaling large models.

Having established the conceptual motivation and benefits of VWN, we now describe its internal mechanism. The input to Virtual Width Networks (VWN) is a widened token embedding, which we refer to as the Over‑Width Embedding. Within VWN, the intermediate representations are correspondingly referred to as Over‑Width Hidden States. To process these states, we replace the standard residual connections with Generalized Hyper‑Connections (GHC)—a more general formulation that unifies the ideas of Hyper‑Connections (HC)~\cite{zhu2024hyper} and Frac‑Connections (FC)~\cite{zhu2025frac}. GHC introduces a flexible mechanism that, with lightweight computation, compresses the Over‑Width Hidden States to the backbone width before feeding them into the attention or feed‑forward modules, and then expands the module outputs back to the Over‑Width width to update the Over‑Width Hidden States for the next layer. Finally, a reduce operator, such as a linear projection, maps the last Over‑Width Hidden States back to the original hidden width before the unembedding layer to produce the output logits.

To better exploit the widened representations, we pair VWN with multi-token prediction (MTP), optimizing both the standard next-token objective and an auxiliary $n$-gram loss.
Intuitively, the denser MTP supervision exercises the expanded virtual space, while the additional representational degrees of freedom from VWN improve short-range compositional modeling, yielding a synergistic effect.

We evaluate VWN across multiple regimes using internal MoE models.
We report training dynamics and token efficiency relative to matched non‑VWN baselines, and assess downstream generalization.
Headline results show that VWN, which expands the embedding width by 8×, achieves the baseline’s next‑token loss with $2.5\times$ fewer tokens and the next‑2‑token loss with $3.5\times$ fewer tokens, with the efficiency advantage increasing as training progresses, as shown in Figure~\ref{fig:174b}.

\paragraph{Contributions.}
Our main contributions are summarized as follows:
\begin{itemize}
    \item \textbf{Virtual Width Networks (VWN).} We introduce VWN, which decouples embedding width from backbone width and enables $r\times$ virtual widening with minimal additional compute through \textit{Generalized Hyper‑Connections} (GHC). Through systematic scaling experiments, we further uncover a log‑linear scaling law between the virtual width factor~$r$ and loss, shedding light on how virtual widening influences model performance.
    
    \item \textbf{Generalized Hyper‑Connections (GHC).} 
We formalize GHC as a unifying formulation that subsumes prior variants (e.g., Hyper‑ and Frac‑Connections) and provides flexible routing and mixing between virtual and backbone hidden states.
    
    \item \textbf{Synergy with Multi‑Token Prediction (MTP).} We demonstrate that VWN synergizes with MTP, yielding consistent improvements in downstream accuracy.
\end{itemize}

\section{Related Works}

\textbf{Scaling Model Capacity.}
Transformer models have demonstrated strong performance gains through increased model width, depth, and data scale~\citep{kaplan2020scaling, hoffmann2022training}. However, increasing hidden dimensionality often leads to quadratic growth in parameters and compute, posing challenges in resource-constrained settings. Several approaches have been proposed to decouple model capacity from computation. For instance, mixture-of-experts (MoE) models~\citep{shazeer2017sparsely, lepikhin2020gshard, fedus2022switch} conditionally activate subnetworks to scale model size efficiently. Our method increases effective capacity while avoiding the quadratic computational cost typically associated with widening hidden dimensions. This is achieved by decoupling the embedding width from the backbone hidden size.

\textbf{Hyper‑ and Frac‑Connections.}
Hyper‑Connections (HC)\citep{zhu2024hyper} and AltUp\citep{baykal2023alternating} enhance model expressiveness by expanding the hidden dimension through low‑cost compositional links across layers. However, in large hidden spaces, HC often under‑utilizes the expanded representations, since each extension is updated using only a few scalar weights, making it difficult to fully exploit the additional capacity. Frac‑Connections (FC)~\citep{zhu2025frac} take the opposite approach: instead of enlarging the hidden size, they partition the existing hidden dimension into multiple smaller segments, thereby realizing HC‑like connectivity without increasing model width. Our proposed \textbf{Generalized Hyper‑Connections (GHC)} integrate the advantages of both—expanding the hidden dimension while further subdividing it into structured sub‑states. This design offers fine‑grained control over capacity usage and enables more efficient utilization of the expanded representational space. Moreover, it introduces additional flexibility: the expansion ratio of the hidden dimension need not be an integer multiple, and such fractional expansions have been empirically validated as effective (see Sec.~\ref{sec:1.5}).

\textbf{Embedding Expansion.}
Recent studies have highlighted the importance of vocabulary scaling in large language models~\citep{tao2024scaling}, showing that expanding the input vocabulary—particularly through hierarchical $n$-gram token embeddings—can systematically improve model expressiveness and training efficiency with negligible computational overhead~\citep{huang2025over}. The Over-Tokenized Transformer framework~\citep{huang2025over} introduced \textit{Over-Encoding} (OE) to scale input representations using multi-gram tokenization and \textit{Over-Decoding} (OD) to enhance output supervision via multi-token prediction objectives. Notably, Multi-Token Prediction (MTP)~\citep{gloeckle2024better} is regarded as an effective instantiation of OD for practical training.

\section{Method}
\label{sec:mehtod}

\begin{figure*}[t]
    \begin{center}
    \includegraphics[width=1\textwidth]{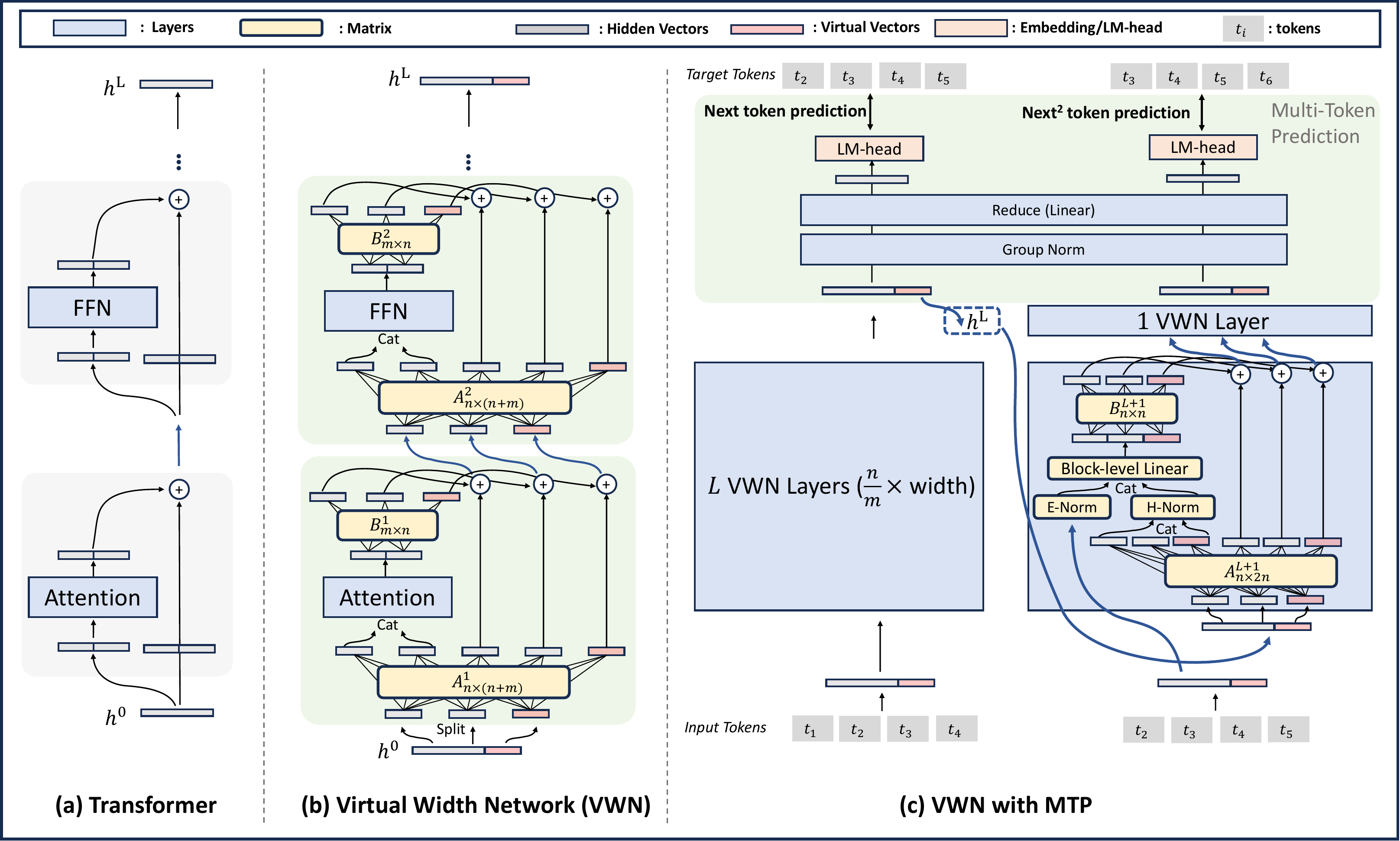}
    \end{center}
  \caption{\textbf{Overview of Virtual Width Networks (VWN).}  
  (a) The standard Transformer maintains a consistent width across input embeddings, intermediate hidden vectors at each layer, and final layer outputs.  
  (b) VWN scales the embedding dimension through \textit{over-width embeddings} while maintaining the layer dimension using lightweight \textit{Generalized Hyper-Connections (GHC)}. These dimensions interact flexibly through small matrices \(\mathbf{A}^l\) and \(\mathbf{B}^l\) ($l$ stands for the layer number).  
  (c) We enable multiple token supervision (\textit{multi-token prediction}), allowing for richer token representations.  
}
    \label{fig:vwn_overview}
\end{figure*}

\subsection{Rethinking the Model Width}

In a standard Transformer model with $L$ layers and model width $D$, the initial token representation $\mathbf{h}^0 \in \mathbb{R}^D$ is obtained through embedding lookup. This representation is subsequently processed through transformer layers, each composed of an attention block and a FeedForward Network (FFN) block. Specifically, at the $l$-th layer, an intermediate hidden vector $\mathbf{h}^l \in \mathbb{R}^D$ is computed from $\mathbf{h}^{l-1}$. The final layer outputs the token representation $\mathbf{h}^{L} \in \mathbb{R}^D$, which is then transformed via a linear head to project it into a $|\mathcal{V}|$-dimensional vocabulary space. The computational complexity for transfomer is $\mathcal{O}(D^2)$, indicating that scaling the model width $D$ results in a quadratic increase in computational cost.

However, the embedding lookup operation represent only a minor fraction of the overall computational cost. Leveraging this insight, we decouple the embedding dimension from the hidden layer dimension so that the embedding dimension to be significantly expanded while maintaining the original hidden dimension for intermediate layer computations. Consequently, this approach preserves nearly the original computational cost while significantly enhancing the representational capacity of token embeddings.

\subsection{Over-Width Embedding}

To increase the embedding dimensions, we propose the \textit{Over-Width Embedding} technique. Given a fixed hidden size $D$, we enlarge the embedding dimension of input to a wider dimension $D'$, resulting in richer token embeddings without a substantial increase in computational overhead.

Formally, let $\mathbf{h}^l \in \mathbb{R}^{D}$ represent the hidden state at $l$-th layer. We partition this hidden vector evenly into $m$ disjoint segments:
\begin{equation}
\mathbf{h}^l = 
\begin{pmatrix} 
{\mathbf{h}^l}_1^\intercal & {\mathbf{h}^l}_2^\intercal & \dots & {\mathbf{h}^l}_m^\intercal 
\end{pmatrix}^\intercal, 
\quad \text{where } {\mathbf{h}^l}_k \in \mathbb{R}^{D/m},\; k = 1, 2, \dots, m.
\end{equation}

Next, we define an expanded embedding vector $\mathbf{e} \in \mathbb{R}^{D'}$, where $D' = \frac{n}{m}D$, with integers $n > m$:
\begin{equation}
\mathbf{e} = 
\begin{pmatrix} 
\mathbf{e}_1^\intercal & \mathbf{e}_2^\intercal & \dots & \mathbf{e}_n^\intercal 
\end{pmatrix}^\intercal, 
\quad \text{where } \mathbf{e}_k \in \mathbb{R}^{D'/n} \text{ with } D'=\frac{n}{m}D .
\end{equation}

Finally, at the input layer, we set $\mathbf{h}'^0 = \mathbf{e}$, thereby utilizing wider token embeddings. 

When the expansion ratio $\tfrac{n}{m}$ is large, a single linear projection can optionally be used to map the original $1\times$ embedding to the wider dimension:
\begin{equation}
\mathbf{E}_{\text{wide}} = \mathbf{W}_{\text{expand}}\, \mathbf{E}_{\text{base}},
\end{equation}
which is like to applying a low-rank decomposition to a very wide embedding table. In addition, one can adopt input-augmentation strategies~\citep{huang2025over} that inject more information per input than a single isolated token embedding to further enrich the widened representation.

For unembedding, the model needs to map the last over‑width hidden states back to the original hidden width $D$ before the unembedding layer. 
We introduce a \textit{reduce operator} $\mathbf{W}_{\text{reduce}} \in \mathbb{R}^{D \times D'}$ that performs a linear projection from the over‑width dimension $D'$ to the original width $D$:
\begin{equation}
\mathbf{h}^L_{\text{reduce}} = \mathbf{W}_{\text{reduce}}\, \mathbf{h}'^L.
\end{equation}
To stabilize training, normalization is applied before the reduce operator, as shown in Figure~\ref{fig:vwn_overview} (c). 
When the expansion ratio $r=\tfrac{n}{m}$ is large, the over‑width dimension $D'$ may become very large (e.g., an 8$\times$ expansion of a 4096‑dimensional hidden size yields a 32K‑dimensional representation).
Instead of directly normalizing across all $D'$ dimensions, we adopt Group Normalization~\cite{wu2018group}, where the group size equals the original hidden size $D$.

\subsection{Generalized Hyper-Connections}

We propose \textit{Generalized Hyper-Connections (GHC)}, a novel method to effectively leverage wider token embeddings while maintaining the original hidden dimension during intermediate layer computations. Specifically, at each layer $l$, GHC introduces a light transformation matrix $\mathcal{GHC}^l$ that encodes weighted relationships between segments of the original hidden representations and the expanded token embeddings. Formally, this matrix is defined as follows:
\begin{align}
\mathcal{GHC}^l 
&=
\left(
\begin{array}{cc}
\mathbf{0} & \mathbf{B}^l \\[3pt]
\multicolumn{2}{c}{\mathbf{A}^l}
\end{array}
\right)
= 
\left(
\begin{array}{cc}
\mathbf{0} & \mathbf{B}^l \\[3pt]
\overset{\circ}{\mathbf{A}}{}^l & \hat{\mathbf{A}}^l
\end{array}
\right)
= 
\begin{pmatrix}
0         & \cdots & 0         & \beta^l_{1,1} & \cdots & \beta^l_{1,n} \\[1mm]
\vdots    & \ddots & \vdots    & \vdots        & \ddots & \vdots        \\[1mm]
0         & \cdots & 0         & \beta^l_{m,1} & \cdots & \beta^l_{m,n} \\[1mm]
\alpha^l_{1,1} & \cdots & \alpha^l_{1,m} & \alpha^l_{1,m+1} & \cdots & \alpha^l_{1,m+n} \\[1mm]
\vdots    & \ddots & \vdots    & \vdots        & \ddots & \vdots        \\[1mm]
\alpha^l_{n,1} & \cdots & \alpha^l_{n,m} & \alpha^l_{n,m+1} & \cdots & \alpha^l_{n,m+n}
\end{pmatrix}
\in \mathbb{R}^{2m \times (m+n)}.
\label{eq:sghc}
\end{align}

Consider the $l$-th network layer $\mathcal{T}^l$, it integrates self-attention layers or feed-forward networks within transformers. The output of the GHC, denoted as $\mathbf{H'}^{l} = \texttt{Reshape}(\mathbf{h'}^{l}, (n, D'/n))$, represents the Over‑Width Hidden States, and can be formulated as:
\begin{align}
\mathbf{H}^{\prime l} &= \mathcal{GHC}^l(\mathcal{T}^{l}, \mathbf{H}^{\prime l-1}) \notag \\
&= {\mathbf{B}^l}^\intercal \mathcal{T}^l\big({\overset{\circ}{\mathbf{A}}{}^{l\intercal}} \mathbf{H}^{\prime l-1}\big) 
+ {\hat{\mathbf{A}}^{l \intercal}} \mathbf{H}^{\prime l-1}. \label{eq:ghc_recurrent_form}
\end{align}

\paragraph{\textbf{Dynamic Generalized Hyper-Connections (DGHC).}}
To further enhance adaptability in the forward process, we introduce a dynamic extension of the GHC method, termed Dynamic GHC (DGHC), where the transformation matrices are adaptively conditioned on input representations $\mathbf{H}'$:
\begin{equation}
\mathcal{GHC}(\mathbf{H}^{\prime}) = \begin{pmatrix}
\mathbf{0}_{m\times m} & \mathcal{B}(\mathbf{H}^{\prime}) \\
\overset{\circ}{\mathcal{A}}(\mathbf{H}^{\prime}) & \hat{\mathcal{A}}(\mathbf{H}^{\prime})
\end{pmatrix}.
\end{equation}


In practice, we adopt the hybrid strategy from \citet{zhu2024hyper,zhu2025frac}, which integrates both static and dynamic parameters, while making slight adjustments to better fit our VWN framework. The dynamic parameters are generated through a lightweight linear projection network. To ensure training stability, input features are initially normalized. Subsequently, a linear transformation coupled with a $\texttt{tanh}$ activation function is applied. The output is then scaled by a small, learnable matrix and combined with the corresponding static matrix:
\begin{align}
\label{eq:norm}
\overline{\mathbf{H}'} &= \texttt{norm}(\mathbf{H}^{\prime}), \\
\label{eq:B}
\mathcal{B}(\mathbf{H}^{\prime}) &=
\mathbf{S}_\beta \circ \texttt{tanh}\left( \frac{\overline{\mathbf{H}^{\prime}}\mathbf{W}_\beta}{\tau} \right)^{\top} + \mathbf{B}, \\[2pt]
\label{eq:A}
\mathcal{A}(\mathbf{H}^{\prime}) &=
\mathbf{S}_\alpha \circ \texttt{tanh}\left( \frac{\overline{\mathbf{H}^{\prime}}\mathbf{W}_\alpha}{\tau} \right) + \mathbf{A}.
\end{align}

where $\tau=\sqrt{D/m}$, $\mathbf{S}_\beta\in\mathbb{R}^{m\times n}$ and $\mathbf{S}_\alpha\in\mathbb{R}^{n\times(m+n)}$ are learnable scaling matrices initialized to $\mathbf{1}$ (same shapes as $\mathbf{B}$ and $\mathbf{A}$, respectively). Let $d_b := D'/n = D / m$ denote the per-block width and view $\mathbf{H}'$ as an $n\times d_b$ matrix. The projection weights $\mathbf{W}_\beta\in\mathbb{R}^{d_b\times m}$ and $\mathbf{W}_\alpha\in\mathbb{R}^{d_b\times(m+n)}$ are learnable parameters that generate the dynamic coefficients. With these shapes, $\overline{\mathbf{H}'}\mathbf{W}_\beta\in\mathbb{R}^{n\times m}$ and $\overline{\mathbf{H}'}\mathbf{W}_\alpha\in\mathbb{R}^{n\times(m+n)}$; the transpose in Eq.~\eqref{eq:B} makes the former $m\times n$ to match $\mathbf{B}\in\mathbb{R}^{m\times n}$, while Eq.~\eqref{eq:A} already aligns with $\mathbf{A}\in\mathbb{R}^{n\times(m+n)}$.

\paragraph{\textbf{Initialization and Implementation.}}
The dynamic parameters $\mathbf{W}_{\beta}$ and $\mathbf{W}_{\alpha}$ in Eqs.~\eqref{eq:B} and \eqref{eq:A} are initialized to 0, while the static matrices are initialized as follows. It is worth noting that we do not perform any dedicated tuning of the initialization, and thus there remains room for improving learning efficiency.

The static matrix $\mathbf{B} \in \mathbb{R}^{m \times n}$ is initialized with a cyclic pattern:
\begin{equation}
\label{eq:init_B}
\mathbf{B}[i, j] = \begin{cases}
1, & \text{if } i = j \bmod m, \\
0, & \text{otherwise},
\end{cases}
\quad \text{for } i \in \{0, \ldots, m-1\}, \, j \in \{0, \ldots, n-1\}.
\end{equation}

The static matrix $\mathbf{A} \in \mathbb{R}^{n \times n}$ is initialized as a block matrix:
\begin{equation}
\label{eq:init_A}
\mathbf{A} = 
\begin{cases}
\begin{pmatrix}
\mathbf{I}_{m\times m} & \mathbf{I}_{m\times m} & \mathbf{0}_{m\times r}
\end{pmatrix}, & \text{if } n = m, \\[10pt]
\begin{pmatrix}
\mathbf{I}_{m\times m} & \mathbf{I}_{m\times m} & \mathbf{0}_{m\times r} \\
\mathbf{0}_{r\times m} & \mathbf{0}_{r\times m} & \mathbf{I}_{r\times r}
\end{pmatrix}, & \text{if } n > m,
\end{cases}
\quad \text{where } r = n - m.
\end{equation}

The static components $\mathbf{B}$ and $\mathbf{A}$ do not utilize weight decay, whereas the dynamic component does. The implementation details can be found in Appendix~\ref{app:implementation}, while the algorithm is presented in Algorithm~\ref{alg:vwn}.

\begin{algorithm}[h]
\caption{Virtual Width Networks (VWN) Forward Pass}\label{alg:vwn}
\begin{algorithmic}[1]
\Require Over-width token embedding $\mathbf{e} \in \mathbb{R}^{D'}$
\Require Fraction rate $m$, Expanded width $n$, Backbone dimension $D$
\Require Network layers $\{\mathcal{T}^1, \ldots, \mathcal{T}^L\}$ and routing matrices $\{\mathbf{A}^l, \mathbf{B}^l\}_{l=1}^L$
\Require Compression matrix $\mathbf{R} \in \mathbb{R}^{n \times m}$
\Ensure Final output $\mathbf{y}$

\State \textbf{Initialize:}
\State $\mathbf{H'}^0 \gets \texttt{Reshape}(\mathbf{e}, (n, D'/n))^\intercal \in \mathbb{R}^{D'/n \times n}$

\For{$l = 1$ to $L$}
    \State $\mathbf{X}^l \gets {\overset{\circ}{\mathbf{A}}{}^{l}}^\intercal \mathbf{H'}^{l-1}$
    \State $\mathbf{z}^l \gets \mathcal{T}^l(\texttt{Reshape}(\mathbf{X}^l, (D,)))$ \Comment{Input to FFN or Attention layer in Transformer block}
    \State $\mathbf{Z}^l \gets \texttt{Reshape}(\mathbf{z}^l, (m, D/m))^\intercal$
    \State $\mathbf{H'}^l \gets {\mathbf{B}^l}{}^\intercal \mathbf{Z}^l + {\hat{\mathbf{A}}^l}{}^\intercal \mathbf{H'}^{l-1}$ 
\EndFor

\State $\mathbf{h}^L \gets \texttt{Linear}(\texttt{GroupNorm}(\mathbf{H'}^L))$
\State $\mathbf{y} \gets \texttt{Unembedding}(\texttt{Norm}(\mathbf{h}^L))$

\State \Return $\mathbf{y}$
\end{algorithmic}
\end{algorithm}

\subsection{Multi-token Prediction}

As for the output layer, previous research \cite{huang2025over} has demonstrated that Multi-Token Prediction (MTP) serves as an approximation of $k$-gram decoding. Building upon this insight, we leverage MTP to provide fine-grained supervised signals by introducing additional VWN layers atop the backbone model, thus constructing an enhanced prediction head. Specifically, following \citet{deepseekai2025deepseekv3technicalreport}, we concatenate the embedding of the next token with the last-layer embedding of the preceding token, applying a linear projection to generate logits, as illustrated in the upper part of part of Figure~\ref{fig:vwn_overview} (c).

However, adopting a single dense linear that mixes hidden states and embeddings as in \citet{deepseekai2025deepseekv3technicalreport} (i.e., a $2D\!\to\!D$ projection) becomes prohibitively expensive under VWN, where the width is expanded by a factor of $r$. A naive dense mixing would scale to $2rD\!\to\!rD$; for $r{=}8$, both parameters and FLOPs grow substantially and are difficult to afford. To address this, we perform mixing with a block-level linear. We partition the $rD$-dimensional vectors into $n=r\times m$ segments of size $D/m$, and apply the same small linear per segment with shape $(2D/m)\!\to\!(D/m)$. In other words, we fuse the hidden-state and embedding features locally within each segment, sharing the linear projector across all blocks. This preserves the benefits of wider VWN representations while keeping the mixing cost comparable to the $r{=}1$ case.

\subsection{Cost Analysis}

\paragraph{\textbf{Computational Cost.}}
The theoretical computational overhead of VWN is relatively low.  We focus on the dominant computational costs. The normalization operation (e.g. RMSNorm) requires $4\frac{n}{m}D$ FLOPs, per token. Calculating the dynamic parameter $\mathbf{A}$ and $\mathbf{B}$ requires $2\frac{(2m+n)n}{m}D$ FLOPs per token. The width connection incurs a cost of $2\frac{(m+n)n}{m}D$ FLOPs, and the depth connection requires $2nD$ FLOPs.  With modest settings of $m=2$ and $n=3$, the normalization, dynamic parameter calculation, and width connection steps amount to $42D$ FLOPs, while the depth connection requires $6D$ FLOPs.  These computational costs are minor for GPU-based training/inference systems, especially considering the I/O overhead associated with activation memory access, which becomes a bottleneck for VWN. To minimize I/O, the normalization, dynamic parameter calculation, and width connection operations are fused into a single GPU kernel.  Furthermore, the width connection can be fused with the subsequent layer normalization in the transformer layer.  When $m$ is small, VWN adds roughly $\tfrac{n}{m}-1$ times the cost of layer normalization and residual addition due to the over‑width hidden states. This overhead is negligible in such settings, though for larger $m$ its effect varies with the configuration.

\paragraph{\textbf{Memory Cost.}}
During model training, intermediate activations must be stored for backpropagation. VWN introduces additional memory overhead for saving the VWN input activations. However, this can be mitigated through inexpensive recomputation. In a typical training framework like Megatron-LM, each token in a vanilla transformer layer requires $34D$ bytes for activation storage, employing selective activation recomputation \cite{korthikanti2023reducing}. VWN primarily adds the cost of saving the inputs for the $\mathbf{A}$ and $\mathbf{B}$, requiring $2 \times 2 \times (\frac{n}{m} + 1)D$ bytes, given that each number is represented using 2 bytes (16-bit float) and there are two width and depth connections per transformer layer.  While attention and FFN inputs are typically saved for weight gradient computation, they can be efficiently recomputed from the width connection. By saving the input to the $\mathbf{A}$ in the width connection and the input to the $\mathbf{B}$ in the depth connection, the subsequent width connection input can be recomputed at a low cost.  Using a factor $\eta$ to represent the ratio of width connection inputs that are saved, the extra activation memory consumption of VWN for a transformer layer is $4\eta \frac{n}{m}D$ bytes.  With modest settings of $m=2$, $n=3$, and $\eta=0.5$ (saving the width connection input for attention and recomputing it for FFN), the added memory consumption is $3D$ bytes, which is approximately 8.8\% of the memory footprint of the vanilla transformer layer. During model inference, the additional memory overhead arises solely from the extra parameters, a negligible amount compared to other memory consumption.
\section{A Connectivity Perspective}
\label{sec:connectivity}

We reinterpret Virtual Width Networks (VWN) through the lens of connectivity as attention along the depth axis. Consider the stack of layers as a ``depth sequence,'' where each layer index is like a token position and hidden states act as a ``vertical KV cache''. Under this view, common connectivity patterns map to attention-like windows over prior layers: (1) a plain feed-forward stack without residuals corresponds to a sliding window of size~1 (each layer processes only its current input and forgets the previous one); (2) residual connections~\cite{he2016deep} implement a window of size~2 (current input plus the immediately preceding one); and (3) dense connectivity~\citep{ma2023denseformer,huang2017densely,xiao2025muddformer} extends the window size to include all previous layers, allowing each layer to reuse all prior representations. VWN with Generalized Hyper-Connections (GHC) sits in between: it realizes a learned, fixed-cost, linear-attention-like mechanism over depth that scales the accessible depth context.

Formally, let the widened state at layer $l$ be a slot matrix
$\mathbf{H}^{\prime\, l} \in \mathbb{R}^{(D/m)\times n}$ with $n$ slots of size $D/m$, and let $r \coloneqq n/m$ be the width expansion measured in $D$-units. The GHC recurrence with the backbone mapping made explicit is in Eq.~\eqref{eq:ghc_recurrent_form}:
$
\mathbf{H}^{\prime l} = {\mathbf{B}^l}^\intercal \mathcal{T}^l\big({\overset{\circ}{\mathbf{A}}{}^{l\intercal}} \mathbf{H}^{\prime l-1}\big) 
+ {\hat{\mathbf{A}}^{l \intercal}} \mathbf{H}^{\prime l-1},
$
where $\big(\hat{\mathbf{A}}^{\,l}\big)^{\intercal}$ transports/attenuates information stored in the slots (a learned carry/forget operator), and $\big(\mathbf{B}^{\,l}\big)^{\intercal}$ writes the current layer's backbone summary into selected slots. Unrolling Eq.~\eqref{eq:ghc_recurrent_form} explicitly yields

\begin{equation}
\label{eq:depth_unroll_explicit}
\begin{aligned}
\mathbf{H}^{\prime\, l}
&= \sum_{t=0}^{l-1}
\left(\prod_{i=0}^{t-1} \big(\hat{\mathbf{A}}^{l-i}\big)^{\intercal}\right)
\big(\mathbf{B}^{l-t}\big)^{\intercal}
\mathcal{T}^{l-t}\big( \big(\overset{\circ}{\mathbf{A}}^{l-t}\big)^{\intercal}, \mathbf{H}^{\prime (l-t-1)} \big) + \left(\prod_{i=0}^{l-1} \big(\hat{\mathbf{A}}^{\,l-i}\big)^{\intercal}\right)\mathbf{H}^{\prime 0}
\end{aligned}
\end{equation}
with the convention that an empty product equals the identity. Equation~\eqref{eq:depth_unroll_explicit} shows that $\mathbf{H}^{\prime\, l}$ linearly aggregates backbone-transformed features from earlier layers, propagated by the ``carry operator'' $\hat{\mathbf{A}}$ and written via $\mathbf{B}$ at each step---capturing the spirit of linear attention over a compressed depth cache.

\paragraph{Choosing $m$.} The memory budget for storing depth information---measured in $D$-units---is $r{=}n/m$. GHC allocates this budget between per-layer fidelity and the number of layers remembered:
\begin{itemize}
    \item With $m{=}1$, the model stores up to $r$ layers at full $D$-dimensional fidelity (fewer layers, higher bandwidth per layer).
    \item With $m{>}1$, the model stores up to $n{=}rm$ layers, each compressed to $D/m$ dimensions (more layers, lower bandwidth per layer).
\end{itemize}
Thus, $m$ controls per-layer compression, $n$ controls the nominal depth window, and $r$ fixes the total memory budget. The learned, input-dependent routing then provides a soft extension beyond the nominal window via attenuation rather than hard truncation. Intuitively, a larger $m$ expands the effective number of remembered layers at the cost of lower per-layer fidelity. For wider models, the increased representational capacity offers sufficient bandwidth to accommodate a larger $m$. Similarly, deeper networks benefit from larger $m$ since enabling each layer to access longer-range, shallower-layer information can alleviate optimization difficulty and improve gradient flow.

\paragraph{Hard vs.\ soft depth windows.}
\begin{itemize}
    \item Hard routing. If $\hat{\mathbf{A}}^{\,l}$ and $\mathbf{B}^{\,l}$ are near-permutation/binary gates, the update behaves like a fixed-size sliding window over depth. With $m{=}1$, there are $r{=}n$ slots of dimension $D$, so the model can retain the last $r$ layers in full fidelity. With $m{>}1$, there are $n{=}rm$ slots of size $D/m$; each layer's $D$-dimensional state is compressed to $D/m$ and written to one slot, giving a hard window of size $n$ in compressed form.
    \item Soft routing. With real-valued, potentially input-dependent $\hat{\mathbf{A}}^{\,l}$ and $\mathbf{B}^{\,l}$ (Dynamic GHC), information is partially retained and mixed across steps. When the spectral radius of $\big(\hat{\mathbf{A}}^{\,l}\big)^{\intercal}$ is below 1, Eq.~\eqref{eq:depth_unroll_explicit} implies exponentially decayed contributions from preceding layers. The effective depth receptive field can exceed the nominal hard window ($>r$ for $m{=}1$ or $>n$ for $m{>}1$), albeit with progressively attenuated and mixed information.
\end{itemize}

\paragraph{A concrete configuration.}
Consider $(m,n){=}(8,64)$, so $r{=}8$. The model maintains $n{=}64$ slots of width $D/8$. Under hard routing, the current layer can leverage the most recent $64$ layers, each represented at $1/8$ of the original dimensionality. Under soft routing, contributions from layers earlier than $64$ may persist with decay, effectively enlarging the ``depth receptive field''.

\paragraph{On the scope of the attention analogy.}
Our analogy to attention chiefly borrows the KV-cache perspective along depth. It does not imply that inter-layer connections are built via similarity scores or pairwise correlations as in standard self-attention. GHC uses learned (static or input-conditioned) routing matrices to carry, compress, and write information across layers at fixed cost, rather than computing dot-product scores or softmax over layer indices.
\section{Experiments}
\subsection{VWN 1.5×}
\label{sec:1.5}
\begin{figure}[h!]
    \centering
    \begin{minipage}{0.5\textwidth}
        \centering
        \includegraphics[width=\linewidth]{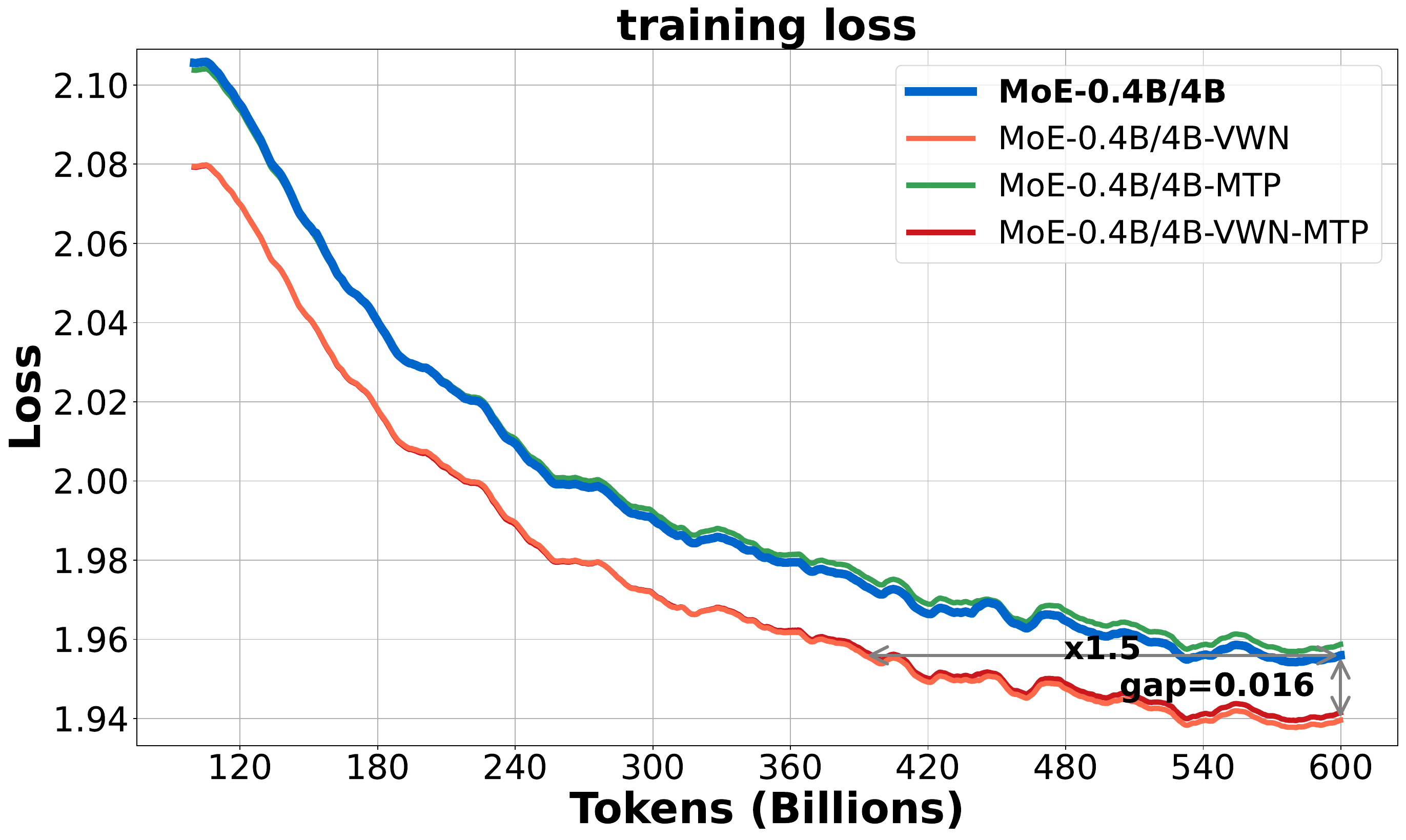}
    \end{minipage}\hfill
    \begin{minipage}{0.5\textwidth}
        \centering
        \includegraphics[width=\linewidth]{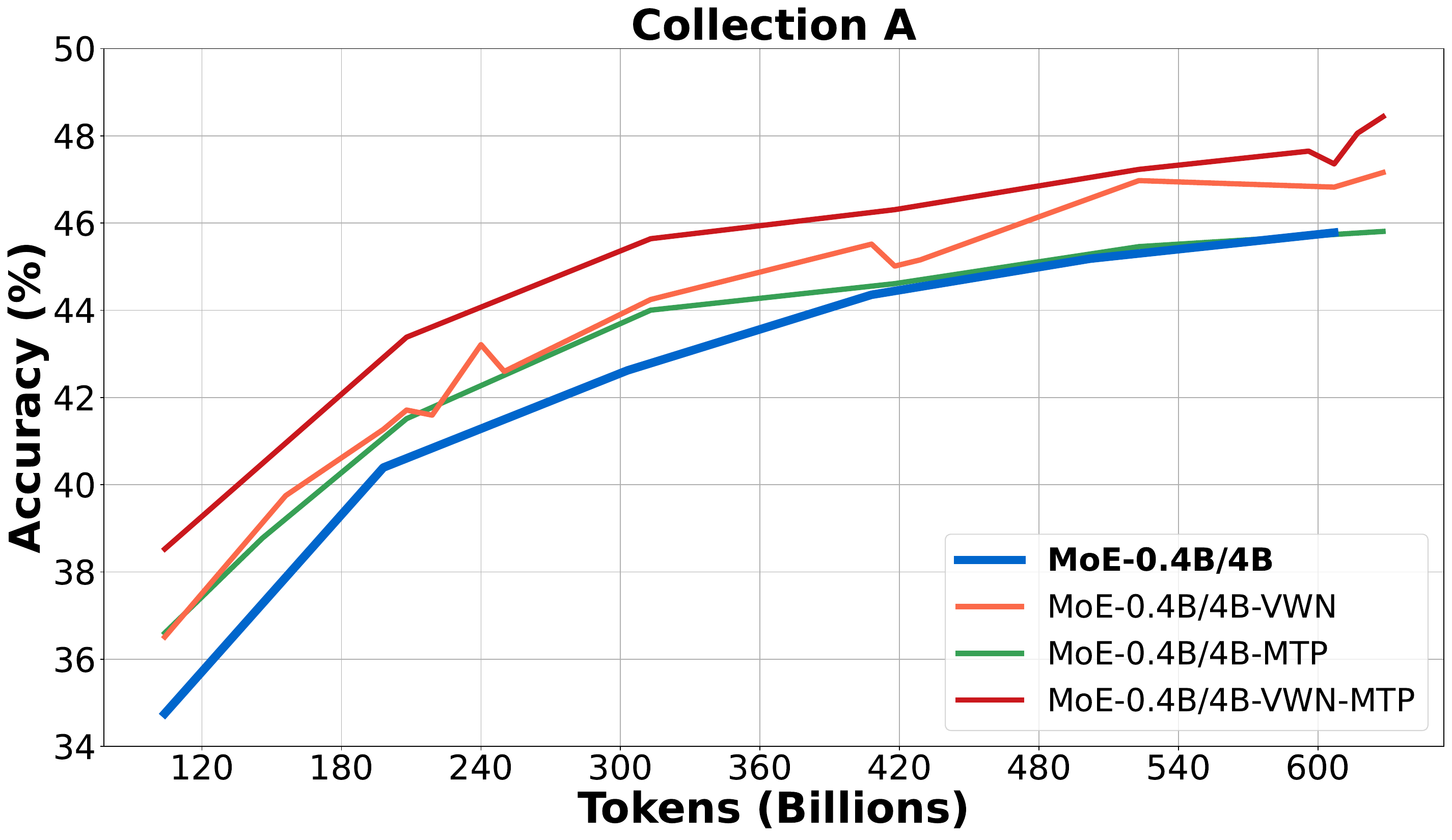}
     \end{minipage}
    \caption{\textbf{Performance of VWN and MTP on 0.4B/4B MoE models.}
\textbf{Left:} Training loss versus seen tokens (billions). VWN lowers the next-token prediction loss, whereas MTP slightly hurts the NTP loss; combining VWN and MTP (VWN+MTP) yields the lowest final loss among the augmented variants but still shows a small gap (~0.016) relative to the baseline metric when MTP is included.
\textbf{Right:} Average downstream accuracy (\%) versus tokens. Both VWN and MTP improve downstream accuracy over the baseline, and their combination delivers the largest gains throughout training. Models: MoE-0.4B/4B (baseline), MoE-0.4B/4B-VWN, MoE-0.4B/4B-MTP, and MoE-0.4B/4B-VWN-MTP.}
    \label{fig:400m}
\end{figure}
To examine the effectiveness of VWN under fractional virtual widening, we use the $1.5\times$ configuration as a representative case. 
We jointly evaluate VWN and Multi‑Token Prediction (MTP) in large‑scale language‑model pre‑training, 
and measure downstream performance on \textbf{Collection A}, defined as the \textit{average score across the benchmarks listed in Table~\ref{tab:benchmarks}}. 
In the $1.5\times$ setting, group normalization preceding the reduce operator (used to aggregate virtual partitions) is omitted.

For our primary evaluation, we conduct comprehensive experiments on internal Mixture‑of‑Experts (MoE~\cite{shazeer2017sparsely}) models of multiple scales, 
including \texttt{MoE 0.4B/4B} and \texttt{MoE 2.5B/30B}, all trained on large‑scale internal datasets. 
Each VWN variant adopts the $(m,n)=(2,3)$ configuration to realize a $1.5\times$ virtual widening relative to the backbone hidden size, 
thereby decoupling the expanded embedding space from the fixed‑width backbone at nearly constant compute. 
This setup enables controlled assessment of VWN and MTP generality across diverse model sizes and realistic production conditions.

\begin{figure}[h!]
    \centering
    \begin{minipage}{0.5\textwidth}
        \centering
        \includegraphics[width=\linewidth]{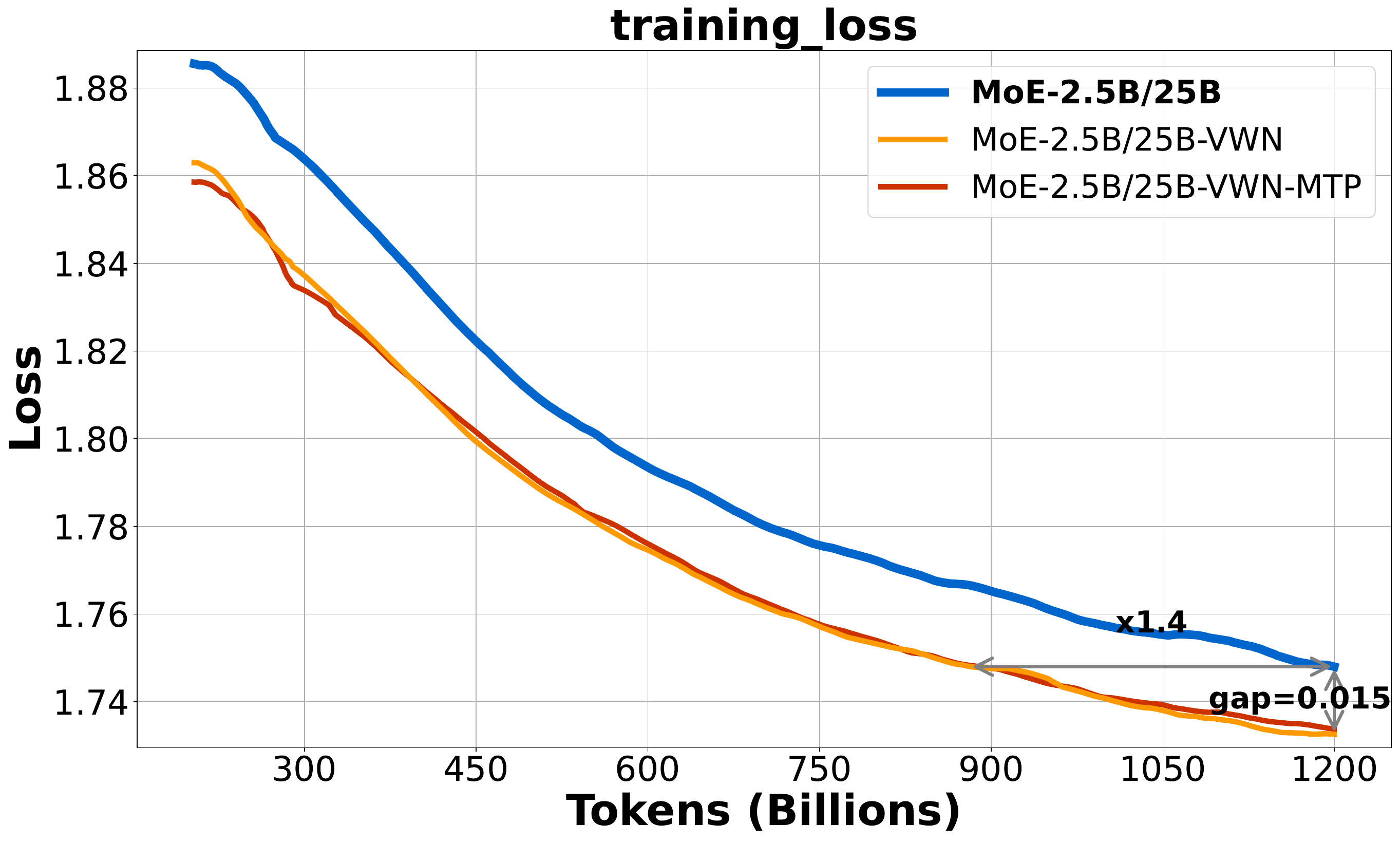}
    \end{minipage}\hfill
    \begin{minipage}{0.5\textwidth}
        \centering
        \includegraphics[width=\linewidth]{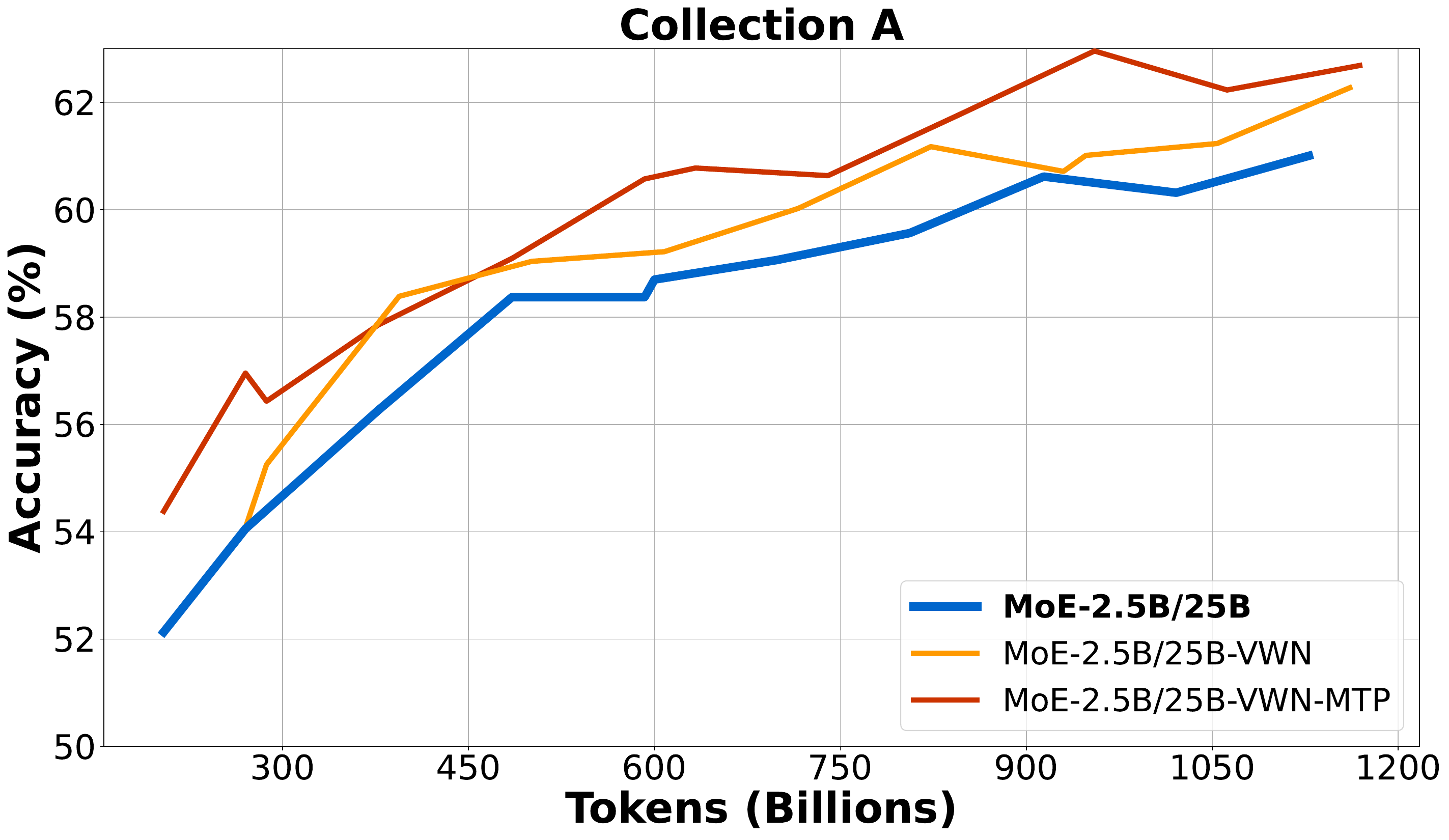}
     \end{minipage}
    \caption{\textbf{Performance of VWN and MTP on 2.5B/25B MoE models.}
\textbf{Left:} Training loss versus seen tokens (billions). VWN reduces the next‑token prediction loss relative to the baseline, and adding MTP on top of VWN does not hurt the loss at this scale, with VWN+MTP reaching the lowest final loss, with a gap of 0.015 versus the baseline at the end of training.
\textbf{Right:} Average downstream accuracy (\%) versus tokens. Both VWN and VWN+MTP outperform the baseline, and VWN+MTP delivers the highest accuracy throughout training. Models: MoE-2.5B/25B (baseline), MoE-2.5B/25B-VWN, and MoE-2.5B/25B-VWN-MTP.}
    \label{fig:2.5b}
\end{figure}
{\bf 0.4B/4B Models.} We study the effects of VWN and MTP on 0.4B/4B MoE models (Figure~\ref{fig:400m}). On the training objective (left), VWN consistently lowers the next‑token prediction (NTP) loss relative to the baseline, whereas MTP slightly increases the NTP loss. The combination VWN+MTP attains the lowest loss among the augmented variants but still shows a gap of 0.016 versus the baseline metric when MTP is included. On downstream evaluation of \textbf{Collection A}, MTP alone is comparable with the baseline, while VWN+MTP delivers the highest gains in average accuracy throughout training.

{\bf 2.5B/25B Models.}
Figure~\ref{fig:2.5b} presents results for 2.5B/25B MoE variants. 
On the training objective (left), VWN reduces the next‑token loss relative to the baseline, 
and adding MTP on top of VWN does not degrade optimization at this scale—both VWN and VWN+MTP achieve similarly low final losses, 
each approximately 0.015 below the baseline. 
On downstream evaluation (right), both variants outperform the baseline, with VWN+MTP consistently yielding the best average accuracy across training.

\subsection{Large Virtual Width}
We study virtual‑width scaling on top of a stronger internal baseline. 
All models include a Multi‑Token Prediction (MTP) head by default, jointly optimizing the standard next‑token and MTP objectives. 
We first run ablations on a 0.8B‑activation MoE (\texttt{MoE‑A0.8B}) to disentangle the effects between increasing $m$ (finer hidden‑partitioning at fixed $r$) and increasing $r$ (greater virtual width at fixed $m$). 
We then scale to a 3.3B‑activation MoE (\texttt{MoE‑A3.3B}) and evaluate the configuration $(m,n)=(8,64)$, corresponding to $r=8$, which delivers an $8\times$ virtual widening of the embedding space while preserving the backbone width. 
We report training dynamics and token efficiency relative to matched non‑VWN baselines. 
Downstream performance is evaluated on \textbf{Collection B}, defined as the \textit{average score across the benchmarks in Table~\ref{tab:benchmarks_b}}.

\begin{figure}[h!]
    \centering
    \begin{minipage}{0.33\textwidth}
        \centering
        \includegraphics[width=\linewidth]{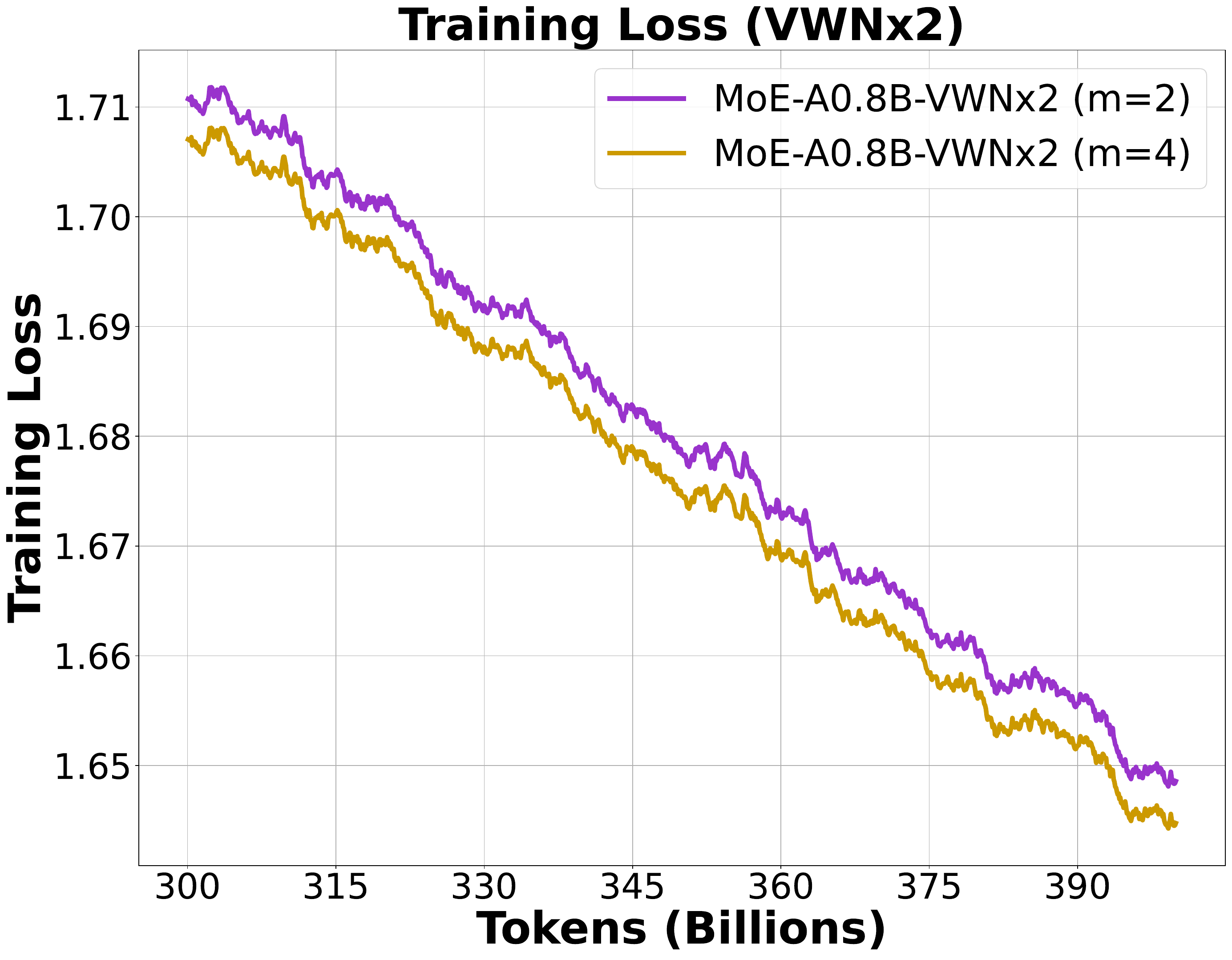}
    \end{minipage}\hfill
    \begin{minipage}{0.33\textwidth}
        \centering
        \includegraphics[width=\linewidth]{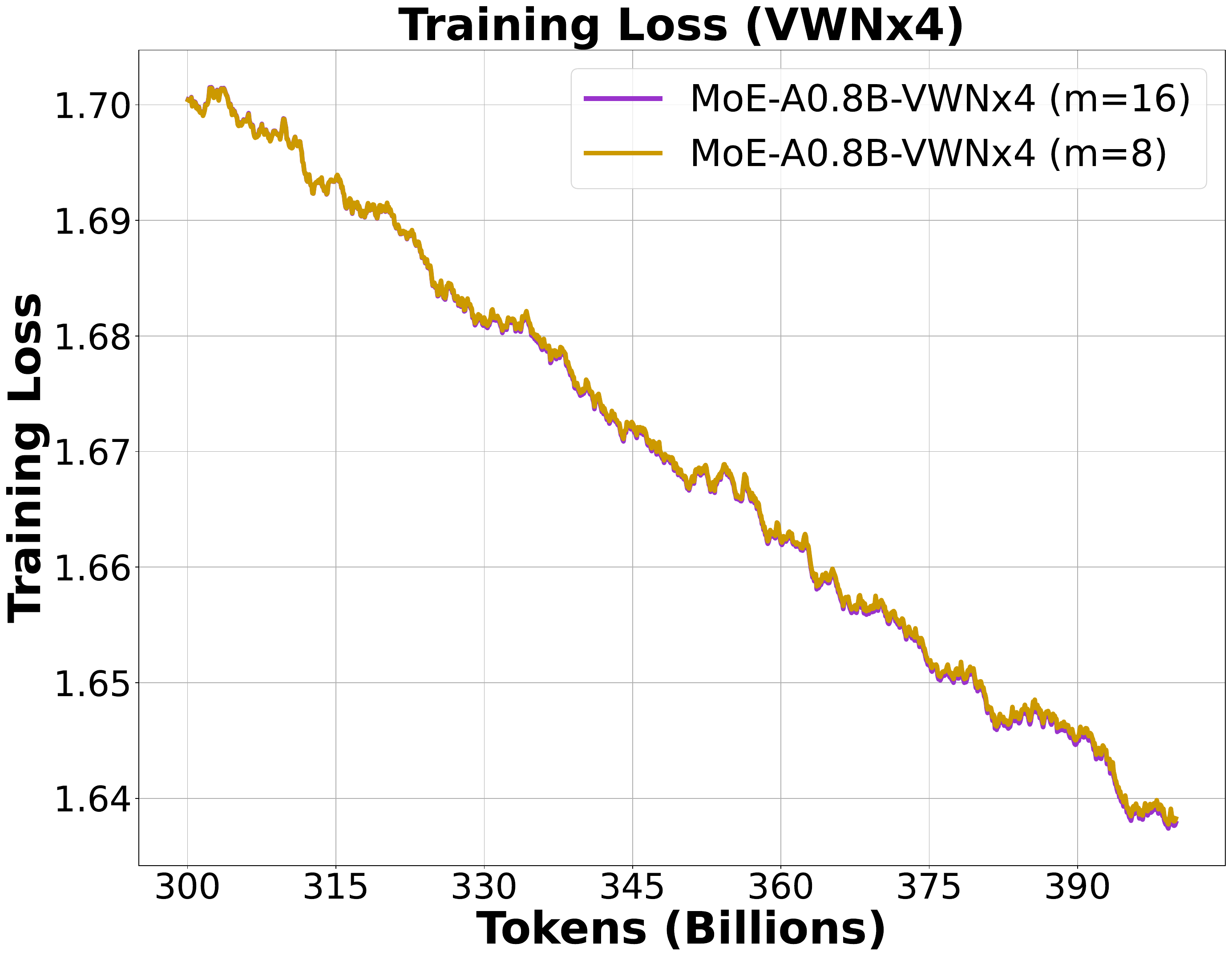}
    \end{minipage}
    \begin{minipage}{0.33\textwidth}
        \centering
        \includegraphics[width=\linewidth]{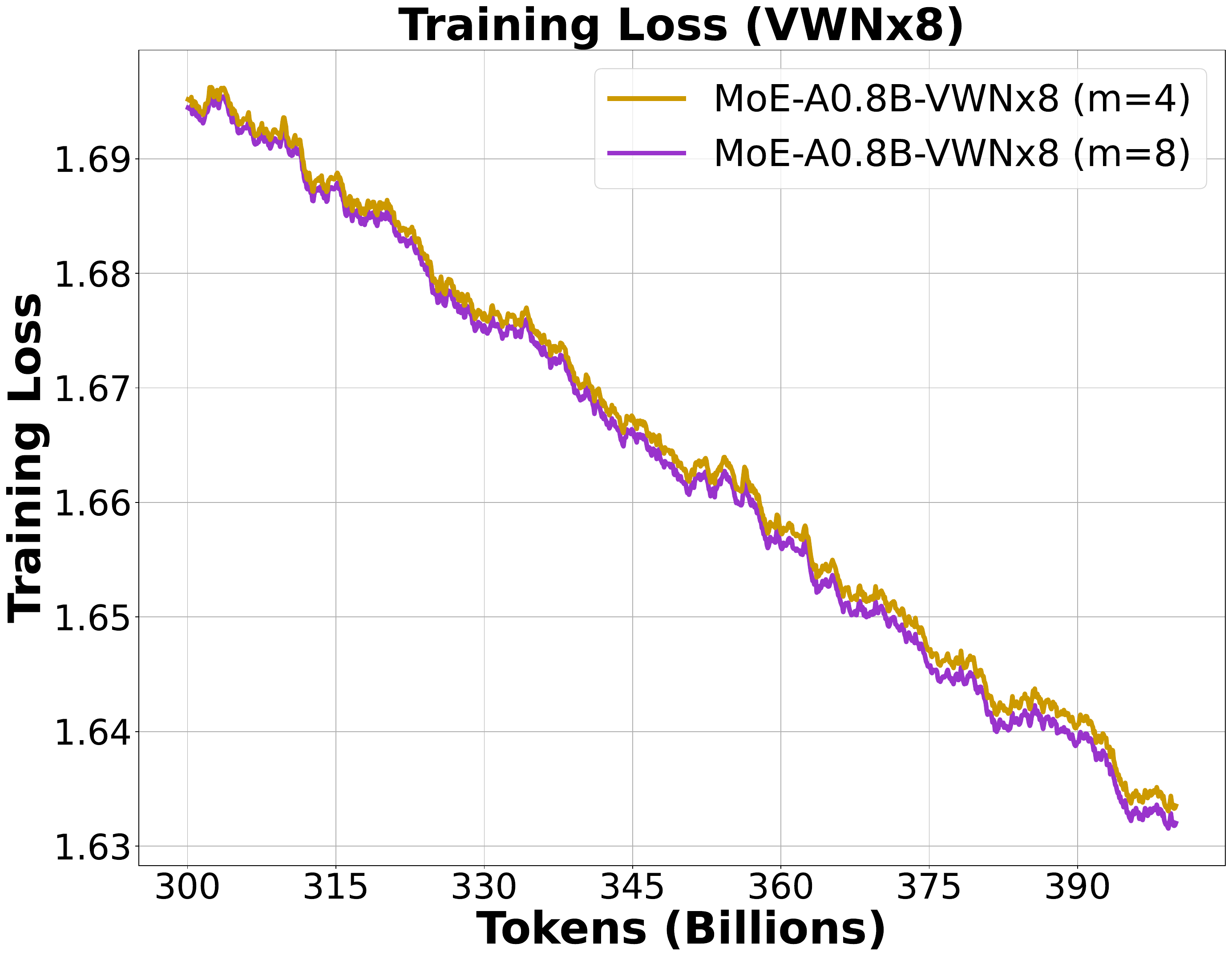}
    \end{minipage}
    \caption{Ablation on the fraction rate $m$ under different virtual‑width factors $r$ on \texttt{MoE‑A0.8B}. Each panel plots next‑token training loss versus seen tokens (billions) for VWN×2 (left), VWN×4 (middle), and VWN×8 (right). At $r{=}2$, increasing $m$ from 2 to 4 produces a modest but visible improvement. When $r{=}4$ or $r{=}8$, varying $m$ between tested values leads to only minor differences, suggesting that beyond $m{\approx}4$ the effect of finer hidden partitioning largely saturates at this model scale.}
    \label{fig:77b_ablation_on_m}
\end{figure}

Figure~\ref{fig:77b_ablation_on_m} presents an ablation on the fraction rate $m$ under different virtual‑width factors $r$ on \texttt{MoE‑A0.8B}. Each plot shows next‑token training loss versus seen tokens (billions). From left to right: $r{=}2$, $4$, and $8$. At $r{=}2$, increasing $m$ from 2 to 4 slightly improves convergence, yielding a noticeable but modest gap. At $r{=}4$, the variants with $m{=}8$ and $m{=}16$ nearly overlap, indicating negligible sensitivity to fraction rate. At  $r{=}8$, the $m{=}4$ and $m{=}8$ curves are similarly close, with marginal advantage for $m{=}8$. Overall, the effect of $m$ diminishes once $m{>}4$, suggesting that, at this scale, partition granularity beyond 4 provides limited benefit. Consistent with the discussion in \S~\ref{sec:connectivity}, we hypothesize that, under a fixed $r$, larger models tend to require higher $m$ to maintain sufficient virtual capacity, whereas smaller models saturate at relatively low $m$ values.

\subsubsection{Scaling Law of the Virtual Width Factor.}
\begin{figure}[h!]
    \centering
    \begin{minipage}{0.33\textwidth}
        \centering
        \includegraphics[width=\linewidth]{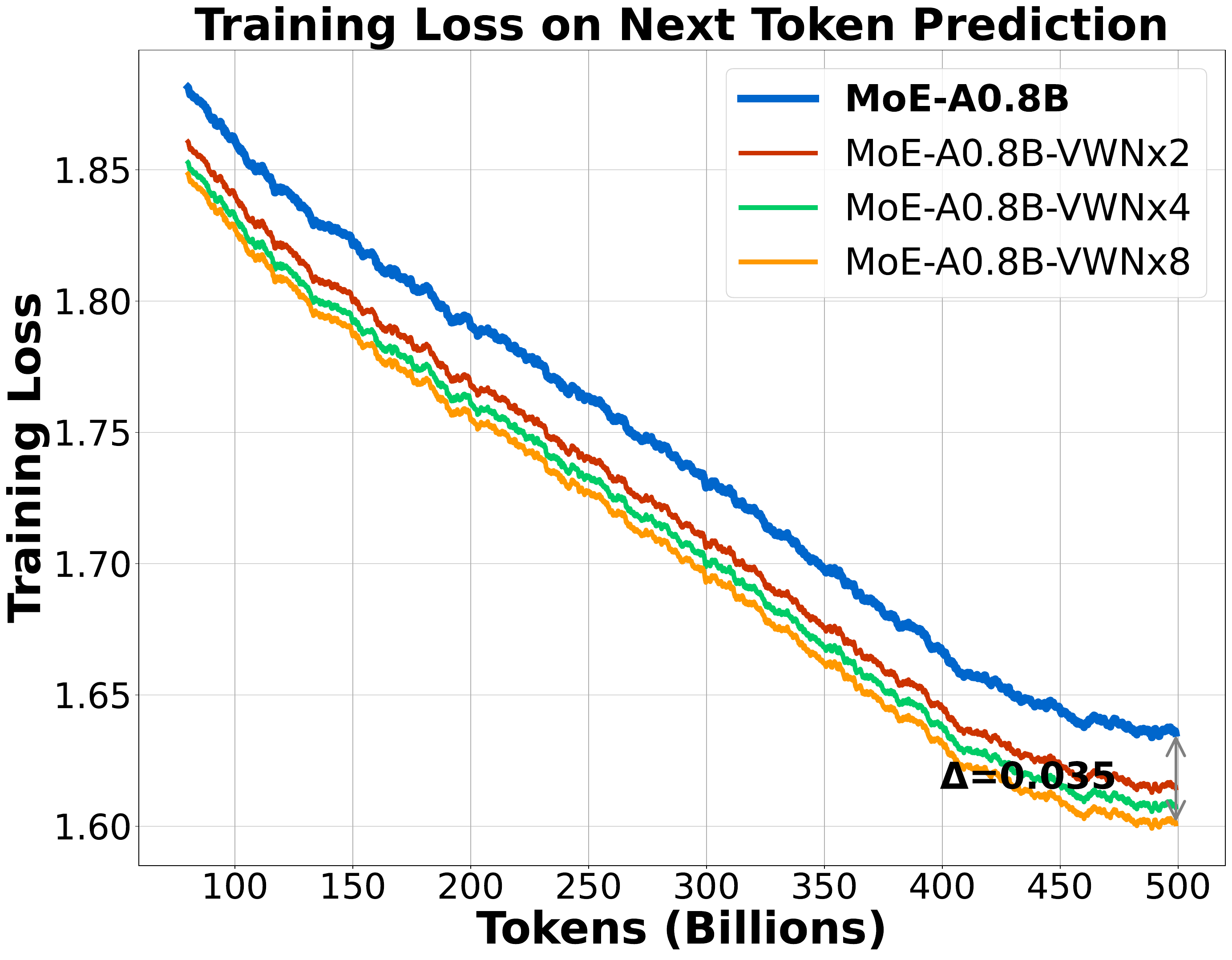}
    \end{minipage}\hfill
    \begin{minipage}{0.33\textwidth}
        \centering
        \includegraphics[width=\linewidth]{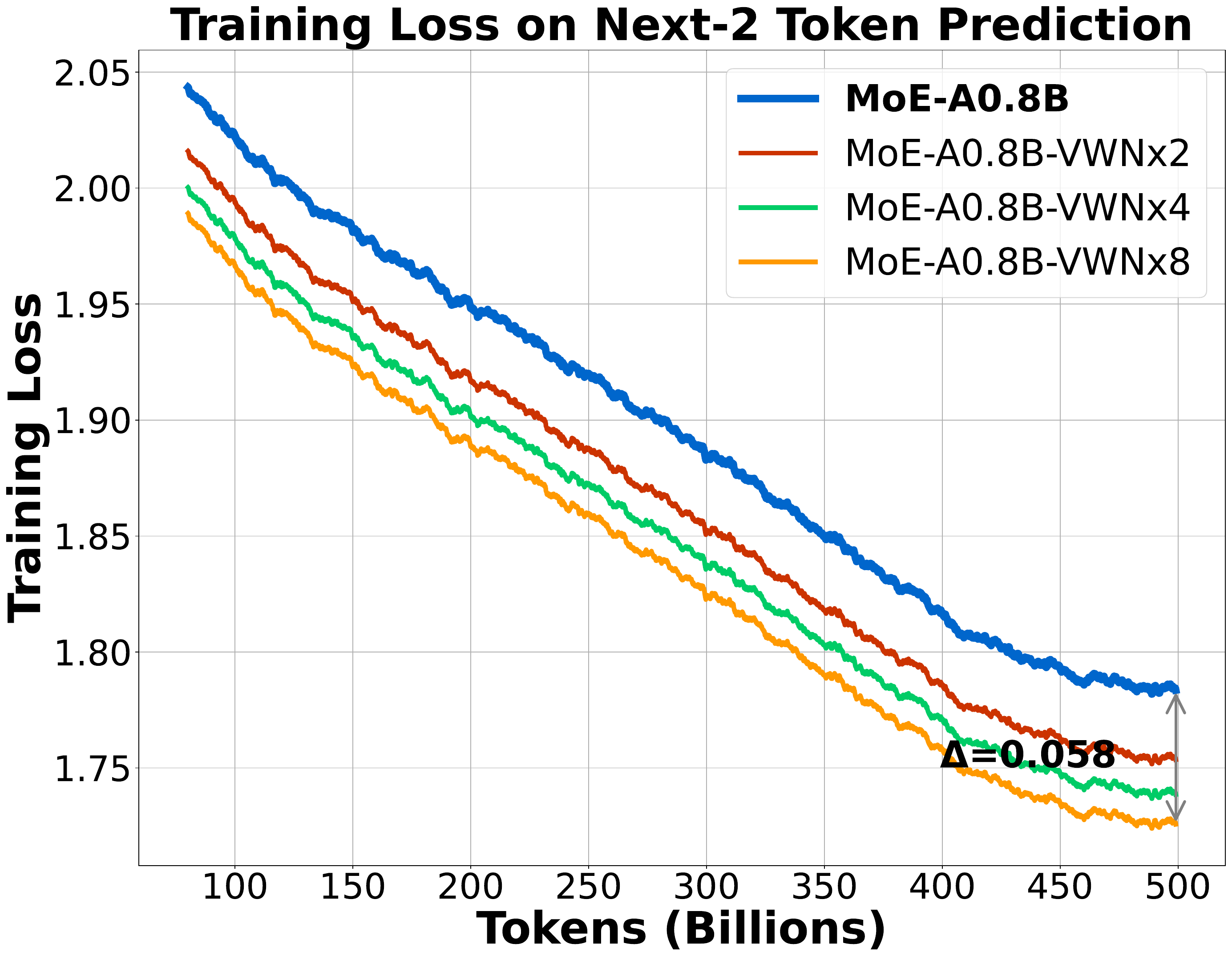}
    \end{minipage}
    \begin{minipage}{0.33\textwidth}
        \centering
        \includegraphics[width=\linewidth]{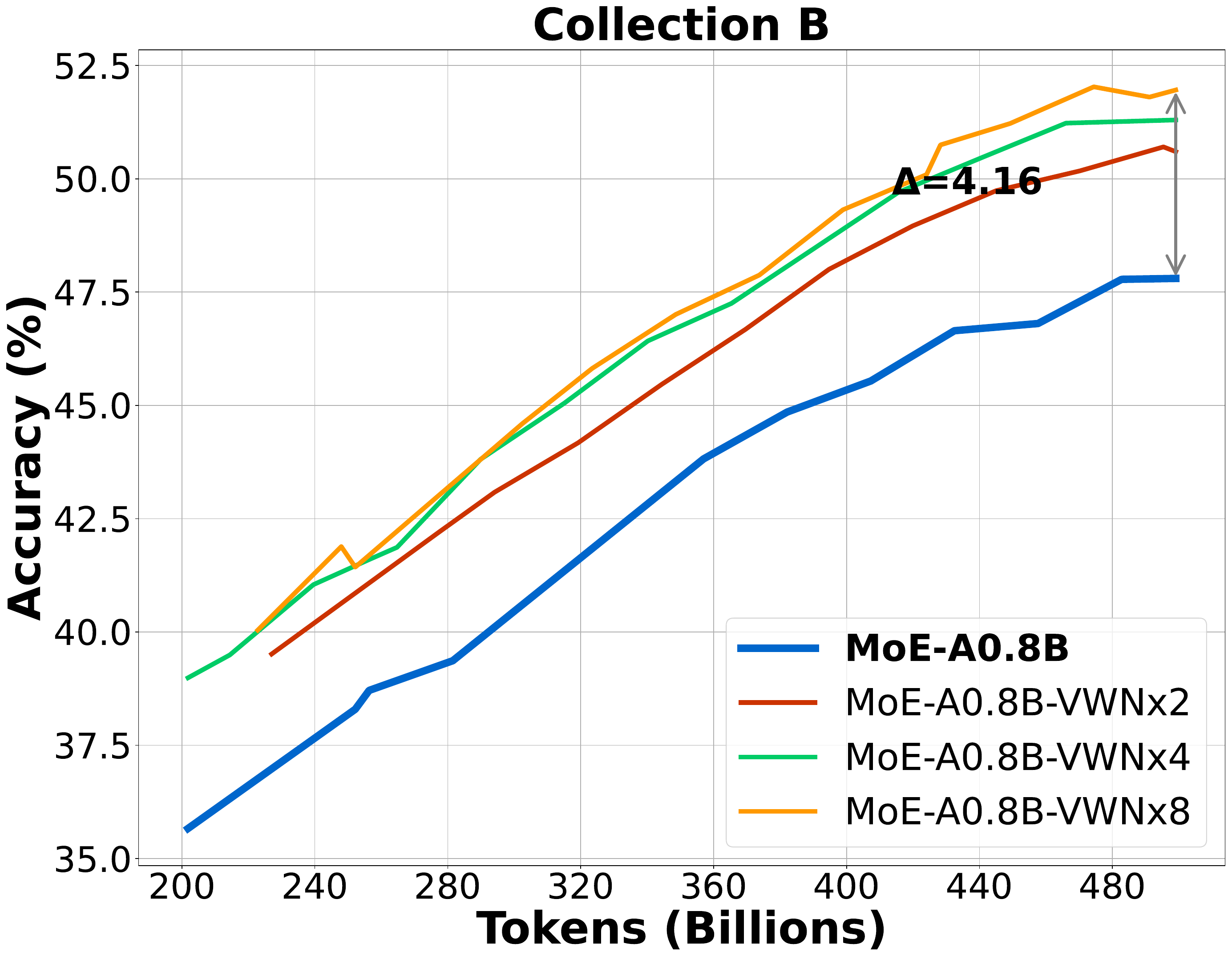}
    \end{minipage}
    \caption{\newtext{Token efficiency of VWN on \texttt{MoE-A0.8B} with a fixed fraction rate $m=8$. 
We vary the virtual width factor by setting $r \in \{2, 4, 8\}$ and $n = r \cdot m = \{16, 32, 64\}$. 
Left/middle: training loss for next-token and next-2-token prediction versus seen tokens. 
Right: average accuracy on \textbf{Collection~B}~\ref{tab:benchmarks_b} versus tokens. 
VWN consistently improves sample efficiency; at 500B tokens, VWN$\times$8 yields $\Delta=0.035$ (next-token loss), $\Delta=0.058$ (next-2 loss), and a $+4.16$-point accuracy gain (\textbf{Collection~B}, Table~\ref{tab:benchmarks_b}) over the non-VWN baseline, by leveraging over-width embeddings and GHC without increasing the backbone width.}}
    \label{fig:77b}
\end{figure}
\begin{figure}[htbp]
    \centering
    \begin{minipage}[t]{0.54\textwidth}
        \vspace{0pt}
        \centering
        \includegraphics[width=\textwidth]{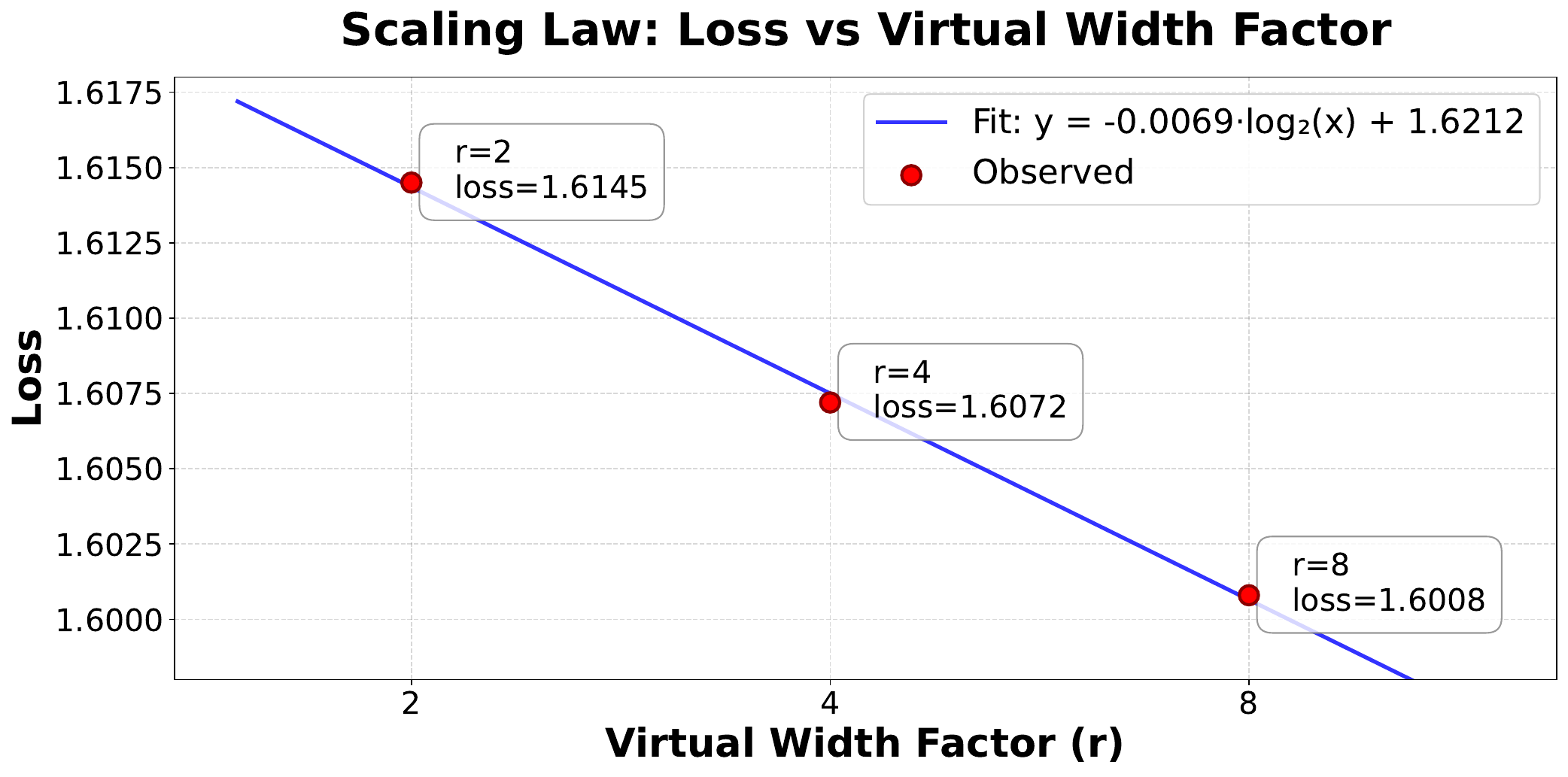}
        \caption{\newtext{Scaling law analysis of the relationship between virtual width factor $r$ and loss. The observed data (red points) are fitted with a log-linear function $y = -0.0069 \cdot \log_{2}(x) + 1.6212$, with a coefficient of determination $R^{2} = 0.9986$.}}
        \label{fig:scaling_law}
    \end{minipage}
    \hfill
    \begin{minipage}[t]{0.44\textwidth}
        \vspace{0pt}
        \centering
        \captionof{table}{\textbf{Scaling behavior of VWN} on \texttt{MoE-A0.8B} at fixed $m=8$. 
    All numbers denote improvements relative to the non‑VWN baseline after training on 500B tokens. Each $\Delta$ value represents the reduction in loss compared to the baseline, 
    and accuracy gains are measured on collection~B.}
        \label{tab:vwn_scaling}
        \vspace{0.5em}
        \begin{tabular}{lccc}
            \toprule
            \textbf{Model} & \makecell{\textbf{$\Delta$ NTP} \\ \textbf{Loss}} & \makecell{\textbf{$\Delta$ Next-2} \\ \textbf{Loss}} & \makecell{\textbf{Acc.} \\ \textbf{(+pts)}} \\
            \midrule
            VWN$\times$2 & 0.020 & 0.030 & +3.20 \\
            VWN$\times$4 & 0.028 & 0.045 & +3.50 \\
            VWN$\times$8 & 0.035 & 0.058 & +4.16 \\
            \bottomrule
        \end{tabular}
\end{minipage}
\end{figure}

We evaluate VWN on \texttt{MoE‑A0.8B} with a fixed fraction rate $m=8$, while varying the virtual‑widening factor $r \in \{2, 4, 8\}$ ($n = r \cdot m$), to analyze how scaling $r$ influences loss and accuracy (Figure.~\ref{fig:77b}). 
Across the 500B‑token training horizon, VWN yields consistent, monotonic gains with larger $r$. 
Table~\ref{tab:vwn_scaling} summarizes improvements over the non‑VWN baseline: 
at 500B tokens, VWN$\times$2, VWN$\times$4, and VWN$\times$8 reduce next‑token loss by $\Delta=0.020$, 0.028, and 0.035,  
next‑2‑token loss by 0.030, 0.045, and 0.058,  
and improve downstream accuracy by +3.2, +3.5, and +4.16  points, respectively.  
The ordering VWN$\times$8 $>$ VWN$\times$4 $>$ VWN$\times$2 $>$ baseline remains consistent throughout training, indicating that enlarging the over‑width embedding at fixed $m$ systematically enhances model capacity. Results of representative benchmark are illustrated in Figure~\ref{fig:77b_obm}. This collection comprises publicly available benchmarks combined using internal task weights, where a 1‑point gain reflects a notable performance difference.

The observed loss reductions follow a log‑linear relation with respect to the virtual‑width factor $r$ (Figure~\ref{fig:scaling_law}). 
A fitted coefficient of $-0.0069$ indicates that each doubling of virtual width corresponds to an approximate loss reduction of 0.0069. 
While the effect size is modest, it suggests a systematic efficiency gain attributable to virtual widening. 
We hypothesize that more expressive backbones and improved mechanisms that more effectively leverage the virtual‑width hidden representations could further amplify the efficiency gains observed with VWN.

\subsubsection{VWN on Large Scale Model} 
As shown in Figure~\ref{fig:174b}, we further evaluate Virtual Width Scaling on a 3.3B‑activation MoE (\texttt{MoE-A3.3B}) using $(m,n)=(8,64)$, where the hidden dimension is divided into $m=8$ partitions, realizing an $8\times$ virtual width expansion. 
To flexibly control the training length, the learning rate is kept constant throughout training.

VWN markedly accelerates optimization. On \texttt{MoE‑A3.3B}, it reaches the baseline’s next‑token loss with 2.5× fewer tokens and the next‑2‑token loss with 3.5× fewer tokens. 
Meanwhile, the next‑token loss gap relative to the baseline increases from $\Delta=0.025$ at early stages to about $\Delta=0.032$ at 3 T tokens, and the next‑2‑token loss gap grows from $\Delta=0.049$ to $\Delta=0.056$. 
These trends indicate that VWN’s advantage amplifies as training proceeds—its relative efficiency not only appears early but also strengthens over time. 
The larger gain on the multi‑token objective further highlights a strong synergy between virtual width and MTP supervision: the over‑width embedding provides richer representational degrees of freedom for short‑range compositional targets, while the Generalized Hyper‑Connections (GHC) transmit gradients between the virtual‑width space and the backbone without expanding intermediate‑layer width. On downstream evaluation across \textbf{Collection B}, VWN achieves a peak average accuracy that is \textbf{+2.16 points higher} than the baseline, confirming that the performance gap persists and continues to widen with extended training.
\section{Conclusion}
We introduced Virtual Width Networks (VWN) as a practical mechanism to decouple representational width from the quadratic compute typically associated with widening. With a modest 1.5× expansion, we observe consistent improvements. When scaling to 8× virtual width, optimization accelerates markedly: next‑token prediction loss converges more than 2× faster and multi‑token prediction loss more than 3× faster relative to the baseline width. Beyond these discrete points, the performance of VWN exhibits a clear scaling behavior. We observe an approximately log‑linear relation between the virtual‑width factor~$r$ and loss reduction, with each doubling of~$r$ corresponding to an average loss decrease of about~0.0069. Although the magnitude of the gain is modest, it suggests that virtual width can be treated as a new and predictable dimension for scaling model efficiency, complementing depth-, width-, and data‑scaling laws in existing literature. VWN integrates cleanly with standard Transformer stacks and training recipes, providing a concrete reference point for studying capacity/compute trade‑offs and for exploring how controlled width expansion can improve quality efficiently.
That said, translating these algorithmic gains into production efficiency depends on systems realities. Despite the promising quality‑per‑compute trade‑off, VWN faces practical constraints: as hidden width grows, communication and memory‑access overheads become non‑negligible, and contemporary hardware is not particularly friendly to very wide activations and cross‑device routing. At present, engineering support for extremely wide configurations remains limited, which constrains deployability. In practice, virtual width expansions in the 1.5×–4× range are more feasible on today’s stacks, while larger expansions may require co‑design of software, memory layouts, and interconnect strategies to fully realize their potential.

\clearpage
\section{Contribution}
\begin{multicols}{2} 
\noindent
\textbf{\color{damaiblue} Contributors} \\
\color{damaiblue} Baisheng Li \\
\color{damaiblue} Banggu Wu \\
\color{damaiblue} Bole Ma \\
\color{damaiblue} Bowen Xiao \\
\color{damaiblue} Chaoyi Zhang \\
\color{damaiblue} Cheng Li \\
\color{damaiblue} Chengyi Wang \\
\color{damaiblue} Chengyin Xu \\
\color{damaiblue} Chi Zhang$^*$ \\
\color{damaiblue} Chong Hu \\
\color{damaiblue} Daoguang Zan \\
\color{damaiblue} Defa Zhu \\
\color{damaiblue} Dongyu Xu \\
\color{damaiblue} Du Li \\
\color{damaiblue} Faming Wu \\
\color{damaiblue} Fan Xia \\
\color{damaiblue} Ge Zhang \\
\color{damaiblue} Guang Shi \\
\color{damaiblue} Haobin Chen \\
\color{damaiblue} Hongyu Zhu \\
\color{damaiblue} Hongzhi Huang \\
\color{damaiblue} Huan Zhou \\
\color{damaiblue} Huanzhang Dou \\
\color{damaiblue} Jianhui Duan \\
\color{damaiblue} Jianqiao Lu \\
\color{damaiblue} Jianyu Jiang \\
\color{damaiblue} Jiayi Xu$^*$ \\
\color{damaiblue} Jiecao Chen \\
\color{damaiblue} Jin Chen \\
\color{damaiblue} Jin Ma \\
\color{damaiblue} Jing Su \\
\color{damaiblue} Jingji Chen \\
\color{damaiblue} Jun Wang \\
\color{damaiblue} Jun Yuan \\
\color{damaiblue} Juncai Liu \\
\color{damaiblue} Jundong Zhou \\
\color{damaiblue} Kai Hua \\
\color{damaiblue} Kai Shen \\
\color{damaiblue} Kai Xiang \\
\color{damaiblue} Kaiyuan Chen \\
\color{damaiblue} Kang Liu \\
\color{damaiblue} Ke Shen \\
\color{damaiblue} Liang Xiang \\
\color{damaiblue} Lin Yan \\
\color{damaiblue} Lishu Luo \\
\color{damaiblue} Mengyao Zhang \\
\color{damaiblue} Ming Ding \\
\color{damaiblue} Mofan Zhang \\
\color{damaiblue} Nianning Liang \\
\color{damaiblue} Peng Li \\
\color{damaiblue} Penghao Huang \\
\color{damaiblue} Pengpeng Mu \\
\color{damaiblue} Qi Huang$^*$ \\
\color{damaiblue} Qianli Ma$^*$ \\
\color{damaiblue} Qiyang Min \\
\color{damaiblue} Qiying Yu \\
\color{damaiblue} Renming Pang \\
\color{damaiblue} Ru Zhang \\
\color{damaiblue} Shen Yan \\
\color{damaiblue} Shen Yan \\
\color{damaiblue} Shixiong Zhao \\
\color{damaiblue} Shuaishuai Cao \\
\color{damaiblue} Shuang Wu \\
\color{damaiblue} Siyan Chen \\
\color{damaiblue} Siyu Li \\
\color{damaiblue} Siyuan Qiao$^*$ \\
\color{damaiblue} Tao Sun \\
\color{damaiblue} Tian Xin \\
\color{damaiblue} Tiantian Fan \\
\color{damaiblue} Ting Huang \\
\color{damaiblue} Ting-Han Fan \\
\color{damaiblue} Wei Jia \\
\color{damaiblue} Wenqiang Zhang \\
\color{damaiblue} Wenxuan Liu \\
\color{damaiblue} Xiangzhong Wu \\
\color{damaiblue} Xiaochen Zuo \\
\color{damaiblue} Xiaoying Jia \\
\color{damaiblue} Ximing Yang \\
\color{damaiblue} Xin Liu \\
\color{damaiblue} Xin Yu \\
\color{damaiblue} Xingyan Bin \\
\color{damaiblue} Xintong Hao \\
\color{damaiblue} Xiongcai Luo \\
\color{damaiblue} Xujing Li \\
\color{damaiblue} Xun Zhou \\
\color{damaiblue} Yanghua Peng \\
\color{damaiblue} Yangrui Chen \\
\color{damaiblue} Yi Lin \\
\color{damaiblue} Yichong Leng \\
\color{damaiblue} Yinghao Li \\
\color{damaiblue} Yingshuan Song \\
\color{damaiblue} Yiyuan Ma \\
\color{damaiblue} Yong Shan \\
\color{damaiblue} Yongan Xiang \\
\color{damaiblue} Yonghui Wu \\
\color{damaiblue} Yongtao Zhang \\
\color{damaiblue} Yongzhen Yao \\
\color{damaiblue} Yu Bao \\
\color{damaiblue} Yuehang Yang \\
\color{damaiblue} Yufeng Yuan$^*$ \\
\color{damaiblue} Yunshui Li \\
\color{damaiblue} Yuqiao Xian \\
\color{damaiblue} Yutao Zeng$^*$ \\
\color{damaiblue} Yuxuan Wang \\
\color{damaiblue} Zehua Hong \\
\color{damaiblue} Zehua Wang \\
\color{damaiblue} Zengzhi Wang \\
\color{damaiblue} Zeyu Yang \\
\color{damaiblue} Zhengqiang Yin \\
\color{damaiblue} Zhenyi Lu$^*$ \\
\color{damaiblue} Zhexi Zhang \\
\color{damaiblue} Zhi Chen \\
\color{damaiblue} Zhi Zhang \\
\color{damaiblue} Zhiqi Lin \\
\color{damaiblue} Zihao Huang \\
\color{damaiblue} Zilin Xu \\
\color{damaiblue} Ziyun Wei \\
\color{damaiblue} Zuo Wang \\
\end{multicols}

Authors are listed in alphabetical order. An asterisk (*) denotes former members of the team.

















\bibliographystyle{plainnat}
\bibliography{main}

\clearpage

\beginappendix

\section{Detailed Downstream Results for \texttt{MoE-A0.8B} Models}
\begin{figure}[h!]
    \centering
    \begin{subfigure}[b]{0.32\textwidth}
        \centering
        \includegraphics[width=\linewidth]{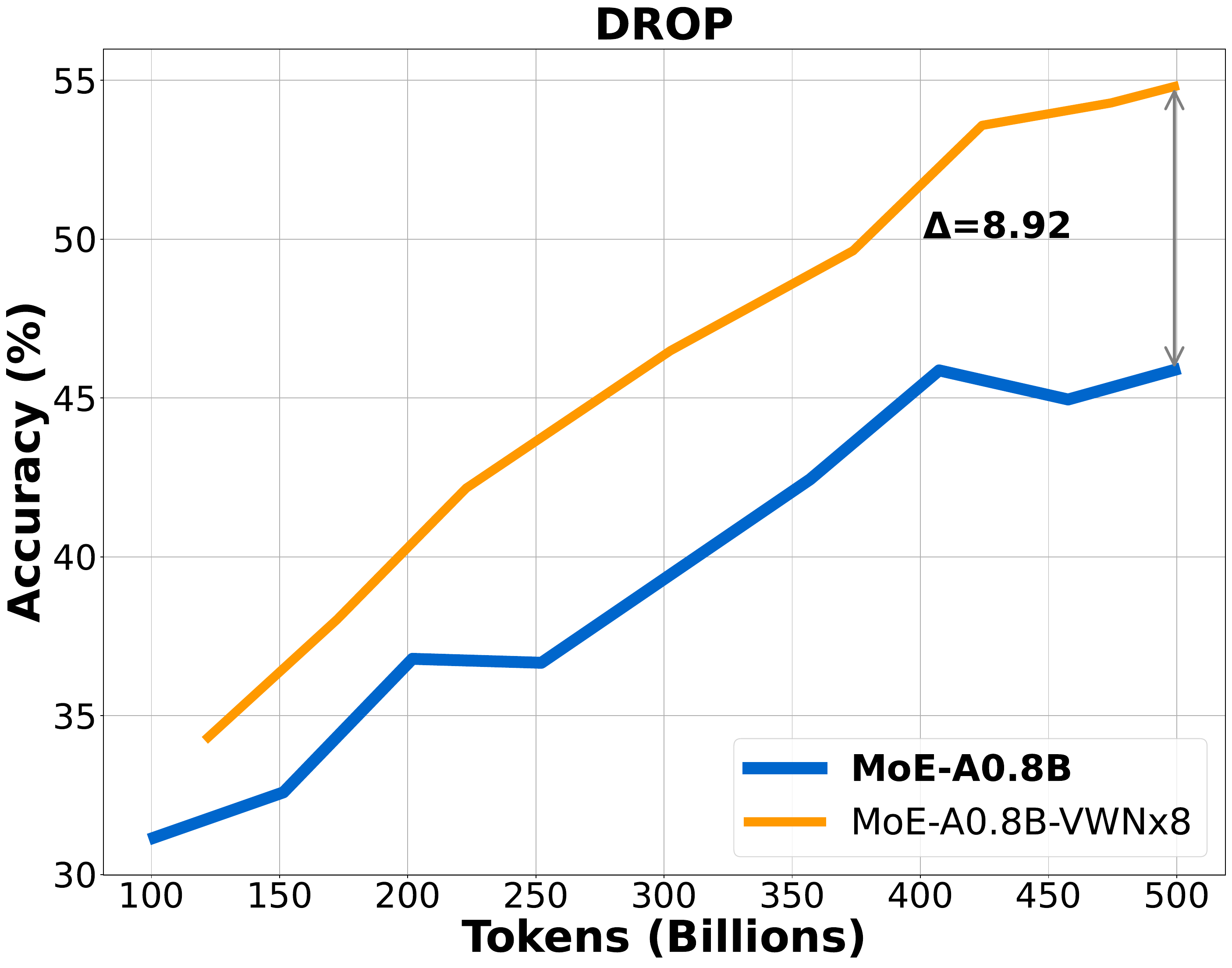}
        \label{fig:77b_training}
    \end{subfigure}\hfill
    \begin{subfigure}[b]{0.32\textwidth}
        \centering
        \includegraphics[width=\linewidth]{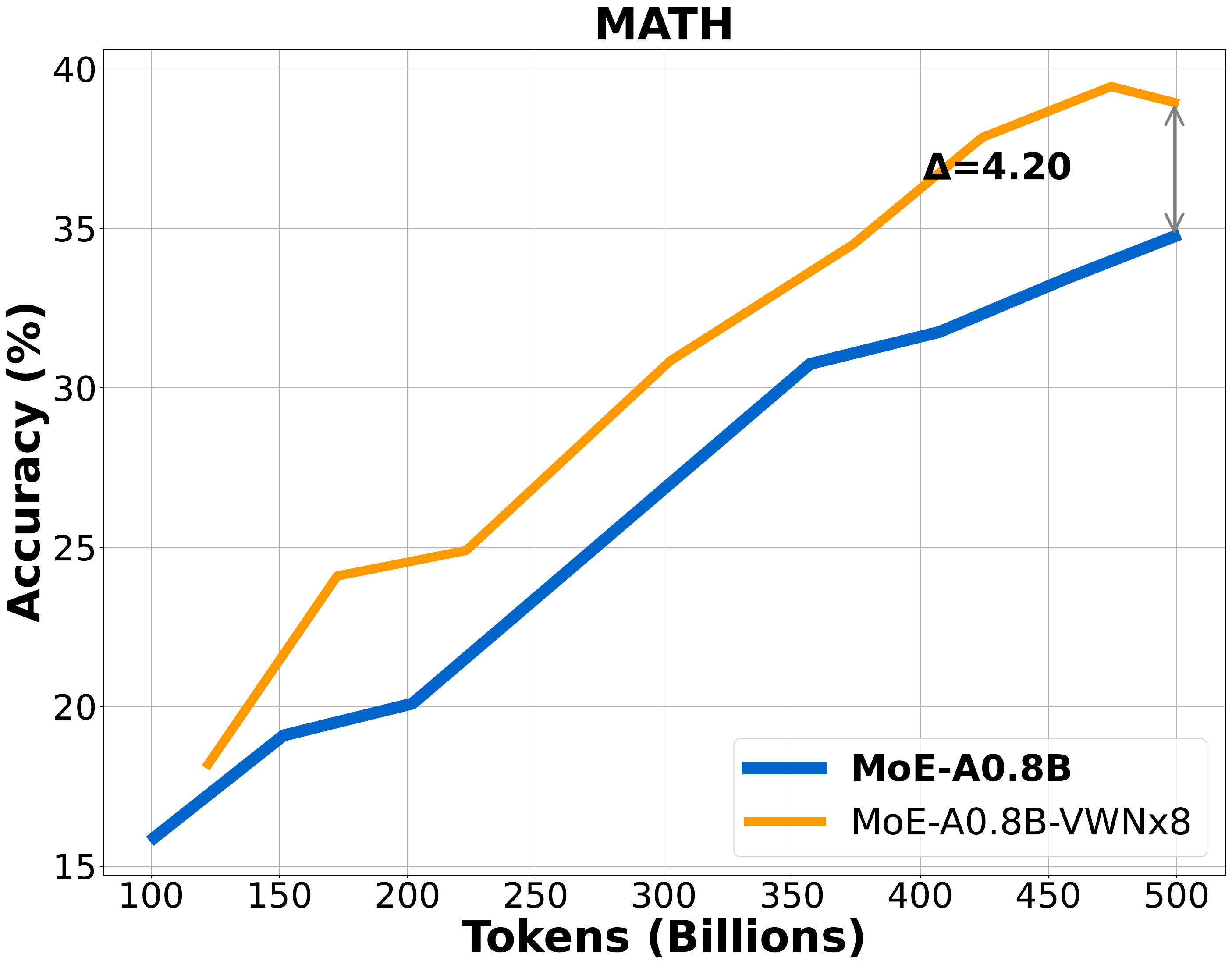}
        \label{fig:77b_downstream}
    \end{subfigure}\hfill
    \begin{subfigure}[b]{0.32\textwidth}
        \centering
        \includegraphics[width=\linewidth]{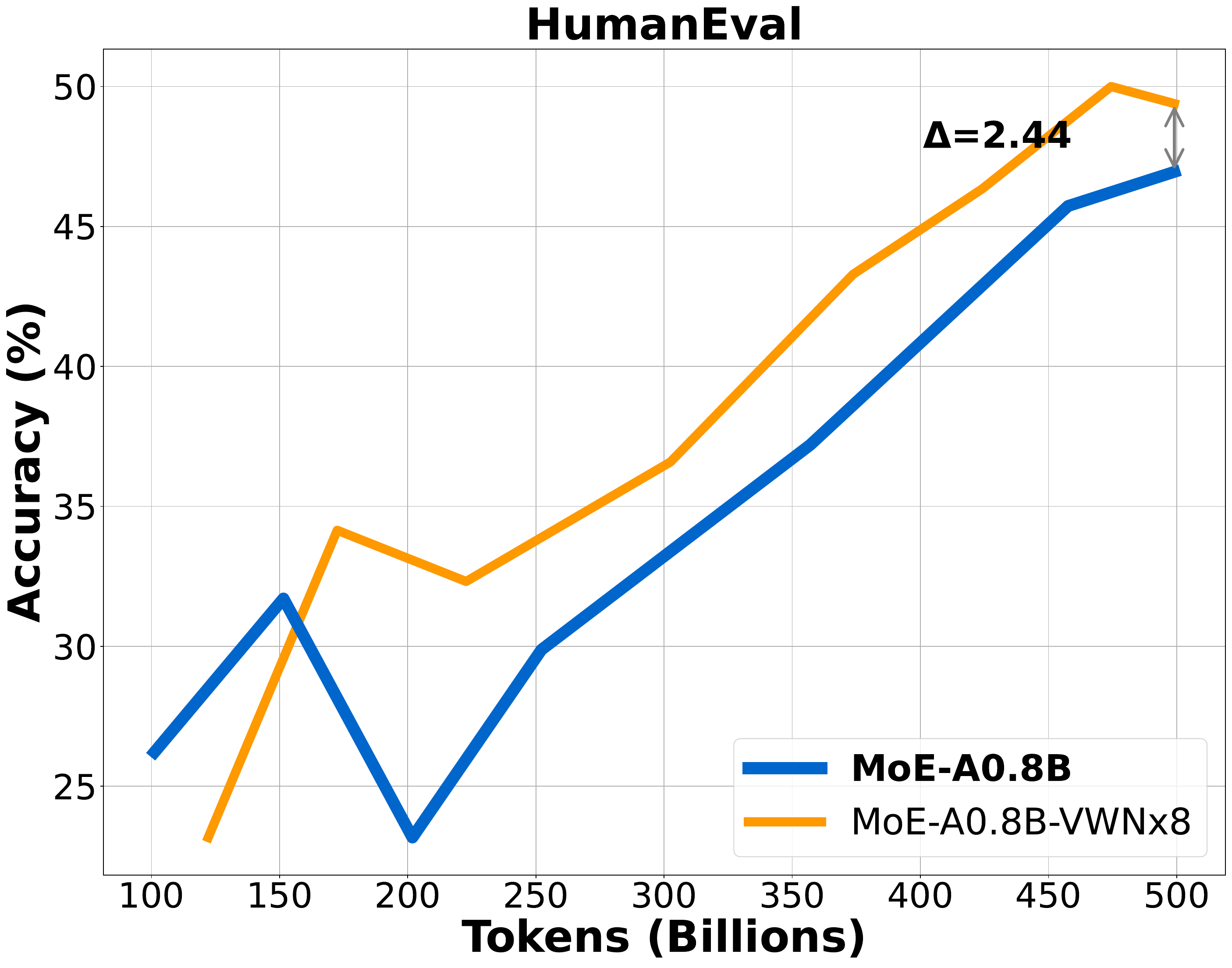}
        \label{fig:77b_next2}
    \end{subfigure}
    
    
    \begin{subfigure}[b]{0.32\textwidth}
        \centering
        \includegraphics[width=\linewidth]{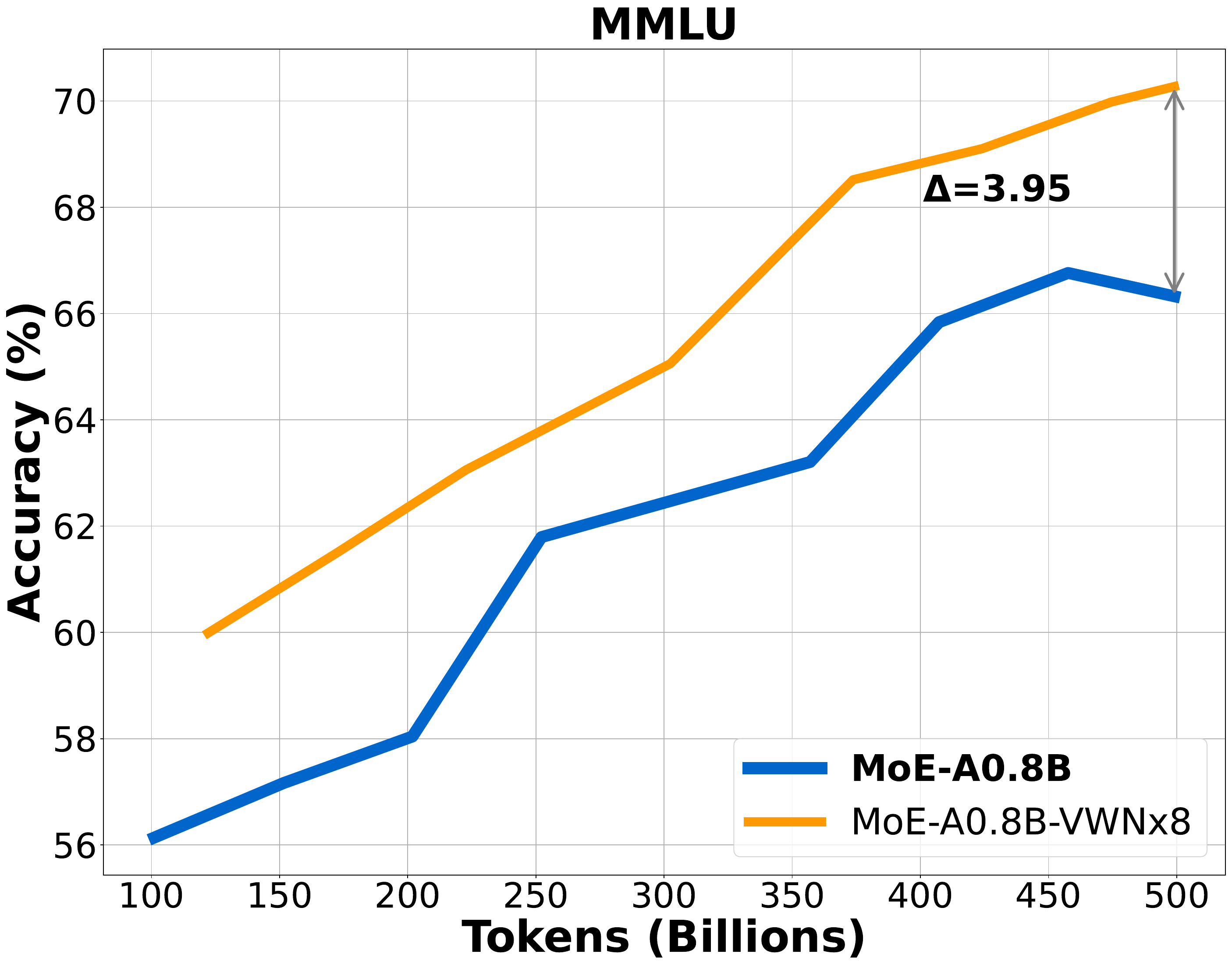}
        \label{fig:image4}
    \end{subfigure}\hfill
    \begin{subfigure}[b]{0.32\textwidth}
        \centering
        \includegraphics[width=\linewidth]{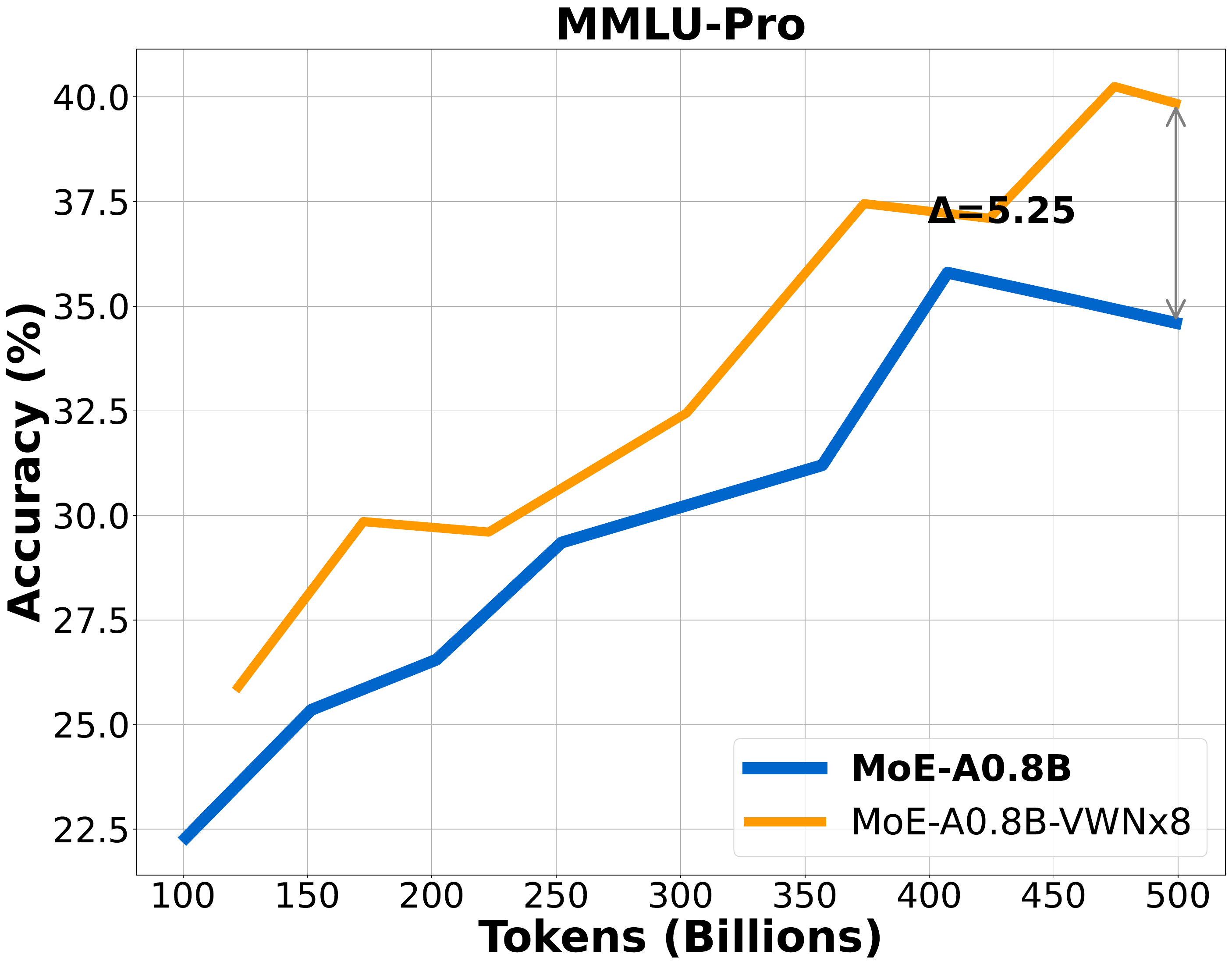}
        \label{fig:image5}
    \end{subfigure}\hfill
    \begin{subfigure}[b]{0.32\textwidth}
        \centering
        \includegraphics[width=\linewidth]{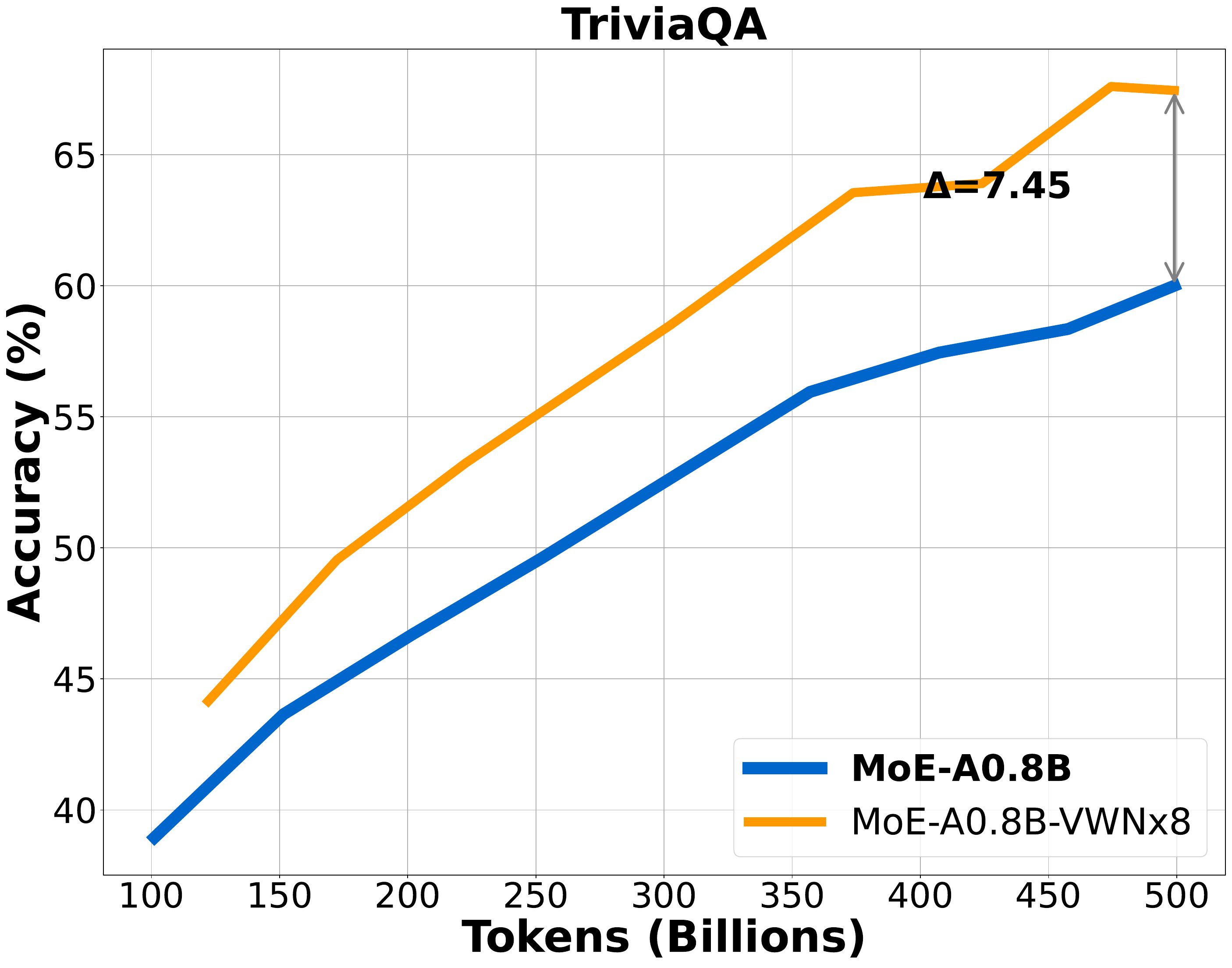}
        \label{fig:image6}
    \end{subfigure}
    
    \caption{\newtext{\textbf{Performance of VWN on MoE-A0.8B across downstream benchmarks.}
        We compare the non‑VWN baseline with \textbf{VWN$\times$8} ($r=8$; $n=r\cdot m=64$).
        \textbf{VWN$\times$8} consistently outperforms the baseline throughout training;
        at 500B tokens it yields +8.92 (DROP), +2.44 (HumanEval), +4.20 (MATH), +3.95 (MMLU),
        +5.25 (MMLU‑Pro), and +7.45 (TriviaQA) accuracy points.}}
    \label{fig:77b_obm}
\end{figure}
\newtext{As shown in Figure~\ref{fig:77b_obm}, which plots token efficiency curves across benchmarks, 
VWN$\times$8 delivers a uniform left‑shift of the learning curves, indicating better sample efficiency on all tasks. 
Improvements are largest on knowledge‑ and reasoning‑heavy benchmarks (DROP, MATH), suggesting that the expanded over‑width embedding improves compositional reasoning and retrieval without increasing core compute. 
HumanEval exhibits smaller gains, consistent with its limited test size. 
The advantages persist late in training, with no regressions observed, indicating that VWN continues to be utilized rather than saturating early. 
Notably, VWN achieves particularly strong gains on tasks with relatively long context, such as DROP and TriviaQA, where modeling extended dependencies and multi‑sentence evidence aggregation benefits most from the enlarged embedding space. 
Overall, VWN consistently transfers its token‑level efficiency gains to diverse downstream domains, strengthening generalization without increasing backbone width.}

\section{Implementation of Generalized Hyper-Connections}
\label{app:implementation}
\begin{algorithm}[H]
\caption{Pseudocode of Generalized Hyper-Connections in a PyTorch-like style.}
\label{alg:torch_fc}
\algcomment{\fontsize{7.2pt}{0em}\selectfont 
}
\definecolor{codeblue}{rgb}{0.25,0.5,0.5}
\lstset{
  backgroundcolor=\color{white},
  basicstyle=\fontsize{7.2pt}{7.2pt}\ttfamily\selectfont,
  columns=fullflexible,
  breaklines=true,
  captionpos=b,
  commentstyle=\fontsize{7.2pt}{7.2pt}\color{codeblue},
  keywordstyle=\fontsize{7.2pt}{7.2pt},
}
\begin{lstlisting}[language=python]
# h: hidden vector (BxLxD)
class GHyperConnection(nn.Module):
    def __init__(self, dim, m, n_in=3, n_out=2):
        super().__init__()
        self.m, self.n_in, self.n_out = m, n_in, n_out
        self.factor = 1.0 / math.sqrt(dim // self.m)
        
        # Initialize static beta: cyclic pattern
        static_beta_tensor = torch.zeros(self.m, n_in)
        for j in range(n_in):
            static_beta_tensor[j % self.m, j] = 1.0
        self.static_beta = nn.Parameter(static_beta_tensor.T.contiguous())

        # Initialize static alpha: block matrix
        init_alpha = torch.cat([torch.eye(self.m), torch.eye(self.m), 
                                torch.zeros((self.m, self.n_in - self.m))], dim=1)
        if self.n_in > self.m:
            part2 = torch.cat([torch.zeros((self.n_in - self.m, self.m * 2)), 
                               torch.eye(self.n_in - self.m)], dim=1)
            init_alpha = torch.cat([init_alpha, part2], dim=0)
        self.static_alpha = nn.Parameter(init_alpha.contiguous())

        # Dynamic parameters
        self.dynamic_alpha_fn = nn.Parameter(torch.zeros((dim // self.m, self.m + self.n_in)))
        self.dynamic_alpha_scale = nn.Parameter(torch.ones_like(self.static_alpha))
        self.dynamic_beta_fn = nn.Parameter(torch.zeros((dim // self.m, self.m)))
        self.dynamic_beta_scale = nn.Parameter(torch.ones_like(self.static_beta))
        self.layer_norm = RMSNorm(hidden_size=dim // self.m)

    def _base_width_connection(self, h, dynamic_fn, dynamic_scale, static_scale):
        h_shape = h.shape
        N, NMM = static_scale.shape
        M = (NMM - N) // 2
        h_reshape = h.reshape((h_shape[:-1].numel(),) + (N, h_shape[-1] // N))
        norm_h = self.layer_norm(h_reshape)
        alpha_beta = (safe_tanh(norm_h @ dynamic_fn.T.to(dtype=norm_h.dtype) * self.factor) 
                      * dynamic_scale[None, ...] + static_scale[None, ...])
        alpha, beta = torch.split(alpha_beta, (M + N, M), dim=-1)
        mix_h = (h_reshape.transpose(1, 2) @ alpha.to(dtype=h_reshape.dtype)).transpose(1, 2)
        return mix_h.reshape(h_shape[:-1] + mix_h.shape[1:]), beta

    def width_connection(self, h):
        dynamic_fn = torch.concat([self.dynamic_alpha_fn.T, self.dynamic_beta_fn.T], dim=0)
        dynamic_scale = torch.concat([self.dynamic_alpha_scale, self.dynamic_beta_scale], 
                                      dim=-1).contiguous()
        static_scale = torch.concat([self.static_alpha, self.static_beta], dim=-1)
        return self._base_width_connection(h, dynamic_fn.to(dtype=h.dtype), 
                                            dynamic_scale.to(dtype=h.dtype), 
                                            static_scale.to(dtype=h.dtype))

    def depth_connection(self, mix_h, h_o, beta):
        h_o_shape = h_o.shape
        h_o = h_o.reshape(h_o_shape[:-1] + (self.m, h_o_shape[-1] // self.m))
        h_i = beta.view(h_o.shape[:2] + beta.shape[1:]).to(dtype=h_o.dtype) @ h_o
        h = h_i + mix_h[..., self.m:, :]
        h_shape = h.shape
        return h.reshape(h_shape[:-2] + (h_shape[-2] * h_shape[-1],)).contiguous()
\end{lstlisting}
\end{algorithm}

\clearpage






\begin{algorithm}[H]
\caption{Pseudocode of transformer with Generalized Hyper-Connections in a PyTorch-like style.}
\label{alg:torch_trans_with_fc}
\algcomment{\fontsize{7.2pt}{0em}\selectfont 
}
\definecolor{codeblue}{rgb}{0.25,0.5,0.5}
\lstset{
  backgroundcolor=\color{white},
  basicstyle=\fontsize{7.2pt}{7.2pt}\ttfamily\selectfont,
  columns=fullflexible,
  breaklines=true,
  captionpos=b,
  commentstyle=\fontsize{7.2pt}{7.2pt}\color{codeblue},
  keywordstyle=\fontsize{7.2pt}{7.2pt},
}
\begin{lstlisting}[language=python]
# h: hidden vector (BxLxD)
# atten_ghyper_connection, ffn_ghyper_connection:  ghyper-connection modules
# attn_norm, ffn_norm: normalization modules

# Attention Block
mix_h, beta = atten_ghyper_connection.width_connection(h)
mix_h_shape = mix_h.shape
h = mix_h[...,:self.rate,:].reshape(mix_h_shape[:-2] + (mix_h_shape[-2] // 2 * mix_h_shape[-1], ))
h = attn_norm(h)
h = self_attention(h)
h = atten_ghyper_connection.depth_connection(mix_h, dropout(h), beta)

# FFN Block
mix_h, beta = ffn_ghyper_connection.width_connection(h)
mix_h_shape = mix_h.shape
h = mix_h[...,:self.rate,:].reshape(mix_h_shape[:-2] + (mix_h_shape[-2] // 2 * mix_h_shape[-1], ))
h = ffn_norm(h)
h = ffn(h)
h = ffn_ghyper_connection.depth_connection(mix_h, dropout(h), beta)

\end{lstlisting}
\end{algorithm}

\clearpage

\section{Downstream Benchmarks}
\begin{table}[H]
\centering
\caption{Downstream Benchmarks Collection A.}
\begin{tabular}{|l|}
\hline
\multicolumn{1}{|c|}{\textbf{Downstream Benchmarks}} \\
\hline
\texttt{ARC\_Challenage}~\citep{allenai:arc} \\
\texttt{BBH}~\citep{suzgun2022challenging} \\
\texttt{DROP}~\citep{dua2019drop} \\
\texttt{WinoGrande}~\citep{sakaguchi2021winogrande} \\
\texttt{Hellaswag}~\citep{zellers2019hellaswag} \\
\texttt{MMLU}~\citep{hendryckstest2021} \\
\texttt{MMLU-Pro}~\citep{wang2024mmlu} \\
\texttt{C-Eval}~\citep{huang2023ceval} \\
\texttt{TriviaQA}~\citep{JoshiTriviaQA2017} \\
\texttt{Ape210K}~\citep{zhao2020ape210k} \\
\texttt{GSM8K}~\citep{cobbe2021gsm8k} \\
\texttt{MATH}~\citep{hendrycksmath2021} \\
\texttt{MBPP}~\citep{austinmbpp2021} \\
\texttt{HumanEval}~\citep{chen2021codex} \\
\texttt{AGIEval}~\citep{zhong2023agieval} \\
\texttt{GPQA}~\citep{rein2024gpqa} \\
\hline
\end{tabular}
\label{tab:benchmarks}
\end{table}

\begin{table}[H]
\centering
\caption{Downstream Benchmarks Collection B.}
\begin{tabular}{|l|}
\hline
\multicolumn{1}{|c|}{\textbf{Downstream Benchmarks}} \\
\hline
\texttt{MMLU}~\citep{hendryckstest2021} \\
\texttt{MMLU-Pro}~\citep{wang2024mmlu} \\
\texttt{C-Eval}~\citep{huang2023ceval} \\
\texttt{AGIEval}~\citep{zhong2023agieval} \\

\texttt{BBH}~\citep{suzgun2022challenging} \\
\texttt{DROP}~\citep{dua2019drop} \\
\texttt{KOR-Bench-Easy}~\citep{ma2024kor} \\

\texttt{MATH}~\citep{hendrycksmath2021} \\

\texttt{MBPP+}~\citep{austinmbpp2021} \\
\texttt{HumanEval}~\citep{chen2021codex} \\
\texttt{McEval}~\citep{chai2024mceval} \\

\texttt{TriviaQA}~\citep{JoshiTriviaQA2017} \\
\texttt{Chinese SimpleQA}~\citep{he2024chinese} \\
\hline
\end{tabular}
\label{tab:benchmarks_b}
\end{table}

\end{document}